\documentclass[10pt]{article}
\usepackage{hmi-preprint}
\usepackage{amsmath}
\usepackage{url}
\usepackage{booktabs}
\usepackage{nicefrac}
\usepackage{microtype}
\usepackage{marvosym}
\usepackage{xcolor}         
\usepackage{tabularx}
\usepackage{multirow}
\usepackage{colortbl}
\usepackage{capt-of}
\definecolor{cvprblue}{rgb}{0.21,0.49,0.74}
\hypersetup{
    colorlinks=true, 
    linkcolor=cvprblue, 
    citecolor=cvprblue, 
    urlcolor=cvprblue,  
}

\DeclareMathOperator{\Propagate}{Propagation}
\DeclareMathOperator{\Project}{Projection}
\newcommand{\appref}[1]{\hyperref[#1]{Appendix~\ref*{#1}}}

\usepackage{wrapfig}
\usepackage{needspace}
\preprintlogos{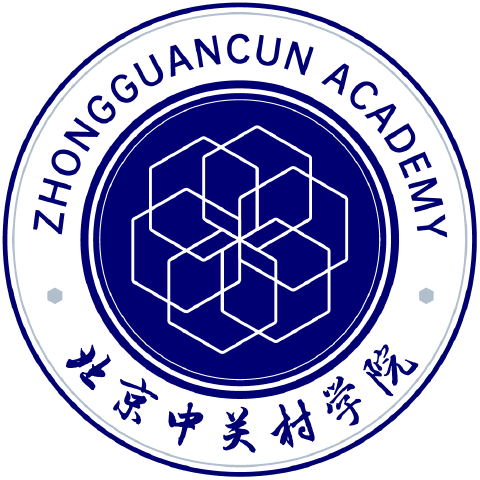}{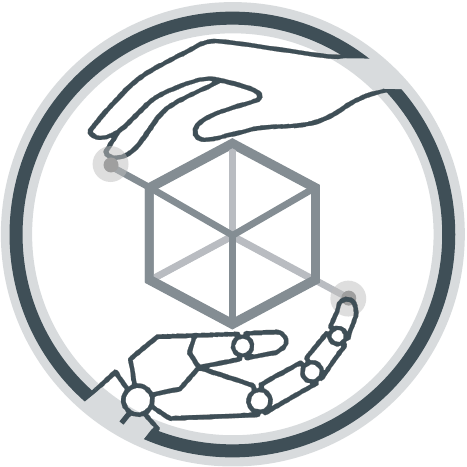}

\preprintdate{October 2026}

\runningtitle{4LC: 4D Latent Cache for Efficient Conditioning in Dynamic World Modeling}

\preprintfooter{\textsuperscript{\Letter} Equal advising.}

\preprintlinkset
  {https://zgca-hmi-lab.github.io/4LC}
  {https://github.com/ZGCA-HMI-Lab/4LC}
  {https://huggingface.co/jianhuizhang/4LC-Video}
  {https://huggingface.co/spaces/jianhuizhang/4LC-Demo}

\preprintteaser{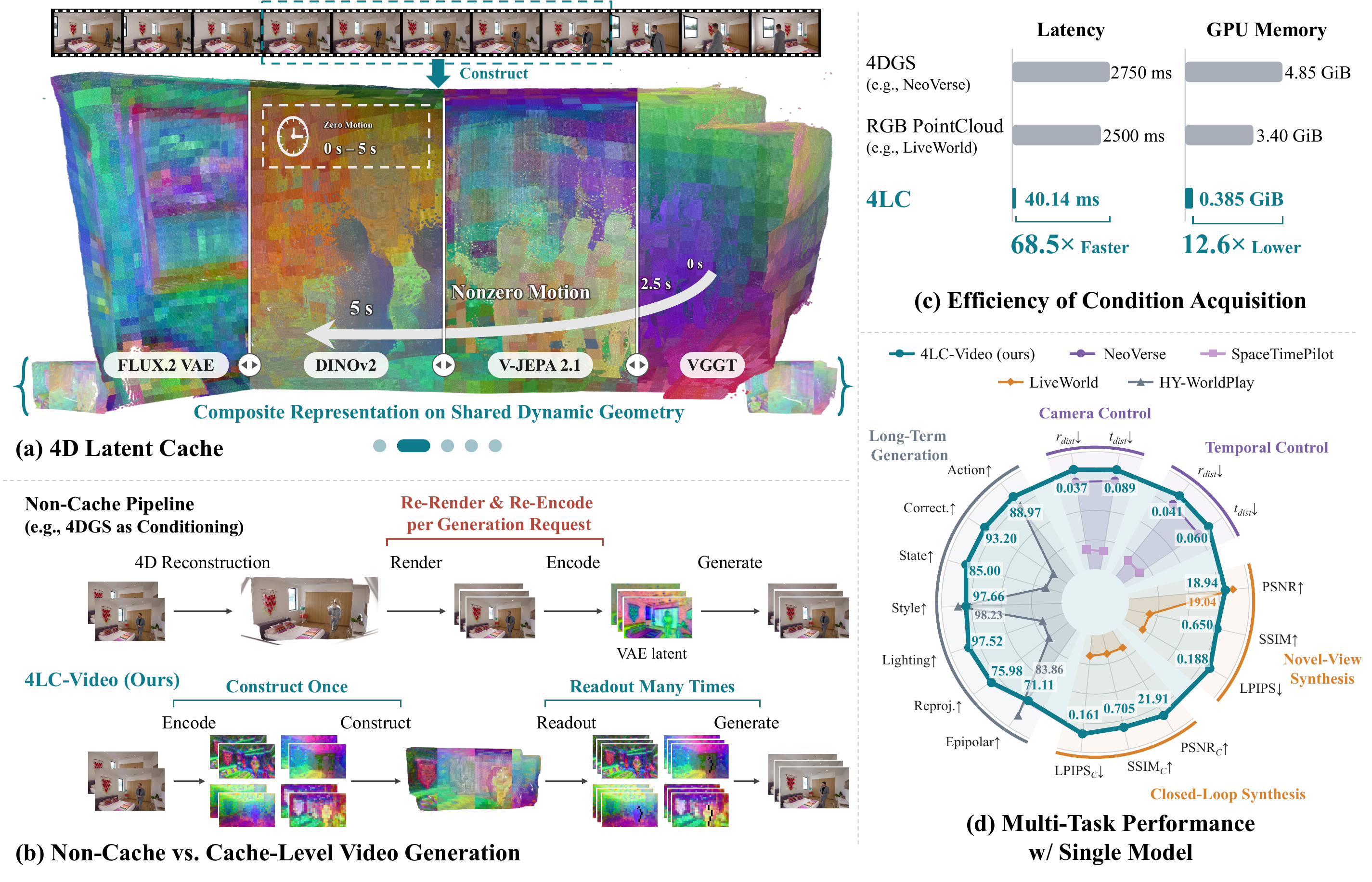}{
    \textbf{Overview of 4LC.} 
    (a) In 4LC, static and dynamic scene content is retained in multiple visual representations for reuse across time and viewpoint requests. 
    (b) Non-cache pipelines render and re-encode a reconstructed scene for each generation request; 4LC encodes source observations once and reads out the cache for subsequent requests. 
    (c) With the same Wan2.1 VAE, 4LC is superior to 4DGS and RGB point clouds in condition-acquisition latency and peak GPU memory. 
    (d) 4LC-Video shows superior performance across five video generation tasks.}

\title{4LC: 4D Latent Cache for Efficient Conditioning in Dynamic World Modeling}

\author{
Jianhui Zhang\textsuperscript{1,2} \hspace{0.12in}
Zuyao Lin\textsuperscript{3,2} \hspace{0.12in}
Pengxin Xu\textsuperscript{4,2} \\[3pt]
Jingdong Wang\textsuperscript{2,\Letter} \hspace{0.12in}
Xingyu Chen\textsuperscript{2,\Letter}
}

\affil{
\textsuperscript{1}Shanghai Jiaotong University \\
\textsuperscript{2} Zhongguancun Academy \\
\textsuperscript{3}Institute of Automation, Chinese Academy of Science \\
\textsuperscript{4}Harbin Institute of Technology
}

\begin{document}

\renewcommand{\topfraction}{0.9}
\renewcommand{\bottomfraction}{0.85}
\renewcommand{\textfraction}{0.08}
\renewcommand{\floatpagefraction}{0.8}
\setlength{\textfloatsep}{12pt plus 2pt minus 2pt}
\setlength{\intextsep}{10pt plus 2pt minus 2pt}
\setlength{\floatsep}{10pt plus 2pt minus 2pt}
\makeatletter
\renewcommand{\paragraph}{\@startsection{paragraph}{4}{\z@}
  {8pt plus 2pt minus 1pt}{-1em}{\normalfont\normalsize\bfseries}}
\setlength{\@fptop}{0pt}
\setlength{\@fpsep}{12pt}
\makeatother

\maketitle

\clearpage
\begin{abstract}
For video generation in dynamic world modeling, observed static and dynamic content should be efficiently reused to provide conditions for repeated generation at different times and viewpoints.
We introduce 4LC, a 4D latent cache with composite visual representations and a flexible spatiotemporal access interface.
As a latent cache, 4LC directly reuses visual representations of static and dynamic content to provide generation conditions, avoiding per-request RGB rendering and re-encoding.
As a composite cache, 4LC binds visual representations from multiple visual encoders (e.g., VAE, DINO, V-JEPA, VGGT) to shared geometry and motion, allowing the composite representations to describe the dynamic world with rich information.
Besides, we design a flexible spatiotemporal interface, which directly reads the cache at independently specified times and viewpoints and combines readouts across observations, providing a common conditioning format for diverse downstream tasks.
Building on 4LC, we develop 4LC-Video to address multiple video generation tasks through different configurations of source observations, spatiotemporal requests, and historical reuse, including camera control, temporal control, novel-view synthesis, scene revisits, and long-term generation.
Our experiments show more efficient condition acquisition, effective spatiotemporal controllability, and scene consistency across tasks using a single 4LC-Video model without task-specific fine-tuning.
\end{abstract}

\section{Introduction}
\label{sec:introduction}

In dynamic world modeling, observed scene information provides conditions for video generation at specified times and viewpoints.
These conditions should convey persistent static structure together with dynamic content at the requested scene time.
Geometry and motion help relate the observed static and dynamic content across viewpoints and scene times~\citep{bib:uni4d,bib:track2view}, supporting consistency in the generated video~\citep{bib:worldreel}.
Because the same observations serve different requests and subsequent generation segments, reusing previously extracted information can reduce the recurring cost of condition acquisition.

Existing conditioning pipelines retain scene information in explicit 3D reconstructions or spatial memories and render it into target views~\citep{bib:gen3c,bib:spatia,bib:long_term_spatial_memory}.
Static and dynamic content can be maintained in separate components~\citep{bib:liveworld,bib:versecrafter} or represented together through shared 4D geometry, as in 4D Gaussian Splatting (4DGS)~\citep{bib:4dgs}.
This shared geometry allows both to be rendered at requested times and viewpoints~\citep{bib:neoverse}.
However, these geometric conditioning pipelines still require RGB rendering and re-encoding for each new generation request (\autoref{fig:teaser}b).
This repeated processing can consume a substantial share of total generation time.
Latent reuse already reduces repeated computation in video world models, both in history processing~\citep{bib:worldkv} and in denoising~\citep{bib:worldcache}.
For conditioning, Mirage caches latent features with scene geometry and warps them directly to target views~\citep{bib:mirage}.
These features lie in the latent space of a specific generator, and dynamic content is excluded from the cache.
These limitations motivate us to investigate a cache mechanism that extends latent reuse to dynamic content, enabling efficient conditioning in dynamic world modeling.

To this end, we introduce 4LC, a 4D latent cache that jointly stores visual representations of static and dynamic content extracted by multiple visual encoders (\autoref{fig:teaser}a).
Motion determines the 3D positions associated with cached representations at the requested scene time, with static content held at its reference position.
These visual representations are bound to the same geometry and motion while retaining their own formats, allowing their complementary information to be used together for conditioning.
4LC provides a spatiotemporal interface that directly reads these representations at independently specified times and viewpoints, combining readouts across observations to supplement uncovered regions.
Once observations are cached, subsequent requests read the stored representations directly instead of rendering and re-encoding them (\autoref{fig:teaser}b), reducing condition-acquisition latency and peak GPU memory (\autoref{fig:teaser}c).

We build 4LC-Video to use these cache readouts as conditions for a shared video generator.
Visual representations of generated observations are appended to the cache as new entries for subsequent segments.
The same model can be configured for camera control, temporal control, novel-view synthesis, scene revisits, and long-term generation by specifying source observations, spatiotemporal requests, and historical reuse.
Across these tasks, a single checkpoint achieves effective spatiotemporal controllability and scene consistency without task-specific fine-tuning (\autoref{fig:teaser}d).

Our main contributions are as follows:
\begin{itemize}
    \item We introduce 4LC, a 4D latent cache with rich visual representations from composite encoders. Its spatiotemporal interface directly reads the cache at independently requested times and viewpoints, without per-request RGB rendering and re-encoding, thereby forming an efficient dynamic cache mechanism for 4D scenes.
    \item We develop 4LC-Video, which includes a spatiotemporal condition adapter to align the rich visual representations of 4LC with a pretrained video diffusion model for downstream video generation.
    \item Without task-specific fine-tuning, a single 4LC-Video model achieves state-of-the-art control accuracy and scene consistency across five generation tasks spanning spatiotemporal control, view synthesis, and long-term generation.
    4LC accelerates condition acquisition by up to $68.5\times$ with $12.6\times$ lower peak GPU memory and reduces its share of end-to-end generation time from 14.9\% to 0.25\%, paving a new route for efficient conditioning and multi-task video generation.
\end{itemize}

\section{Related Work}
\label{sec:related}

\paragraph{Dynamic World Modeling.}
Dynamic world modeling aims at controllable video generation within a persistent scene that evolves over time~\citep{bib:aether,bib:teleworld,bib:inspatio_world,bib:memory_forcing}.
BulletTime~\citep{bib:bullettime} and SpaceTimePilot~\citep{bib:spacetimepilot} control viewpoint and scene time as separate inputs, including frozen and reversed motion.
Spatia~\citep{bib:spatia} maintains consistency across segments by updating a point-cloud memory.
WorldMem~\citep{bib:worldmem} and VMem~\citep{bib:vmem} retrieve past views by field-of-view overlap and surfel indexing, respectively.
Interactive world models support long-term generation over successive segments~\citep{bib:infinite_world,bib:hy_worldplay,bib:lingbot_world,bib:live,bib:yume15}.
WorldKV~\citep{bib:worldkv} retrieves relevant cached history and GIM-World~\citep{bib:gim} compresses past observations into compact memory.
When scene time advances between visits, dynamic content must also remain coherent~\citep{bib:mosaicmem,bib:hydra}.
LiveWorld~\citep{bib:liveworld} simulates out-of-sight dynamics, and VerseCrafter~\citep{bib:versecrafter} separates a static point cloud from per-object trajectories.
To avoid the complexity of maintaining separate components, 4LC binds representations of static and dynamic content to shared geometry and motion, reading them flexibly at independently specified times and viewpoints.

\paragraph{Reconstruction-based Video Generation.}
Reconstruction-based video generation guides video models with renderings of reconstructed geometry.
Several methods condition generation on point-cloud or 3D-cache renderings to complete unseen or poorly reconstructed regions~\citep{bib:gen3c,bib:flexworld}.
For dynamic scenes, 4DGS~\citep{bib:4dgs} and feed-forward 4D reconstruction~\citep{bib:d4rt,bib:vdpm,bib:4rc,bib:4dgt} model geometry and motion, and NeoVerse~\citep{bib:neoverse} renders them at requested times and viewpoints as conditions.
TrajectoryCrafter~\citep{bib:trajectorycrafter} and Vista4D~\citep{bib:vista4d} redirect the camera of monocular videos through point-cloud renderings that preserve source motion.
These pipelines re-encode rendered RGB views for every generation request.
Feature-space approaches instead condition on features carried by reconstructed geometry~\citep{bib:npbg,bib:latentsplat,bib:latent_radiance_fields}.
MVSplat360~\citep{bib:mvsplat360} and Splatent~\citep{bib:splatent} render Gaussian features into diffusion latents for novel-view synthesis.
Mirage~\citep{bib:mirage} warps cached latent features to target views.
MVSplat360, Splatent, and Mirage avoid target RGB encoding, yet their features lie in a specific generator's latent space and cover only static content.
For efficient conditioning, 4LC reads out cached representations directly without per-request re-encoding, and its multi-encoder representations add complementary information.

\section{Method}

\label{sec:4lc}

\begin{figure}[tbp]
    \centering
    \includegraphics[width=\linewidth]{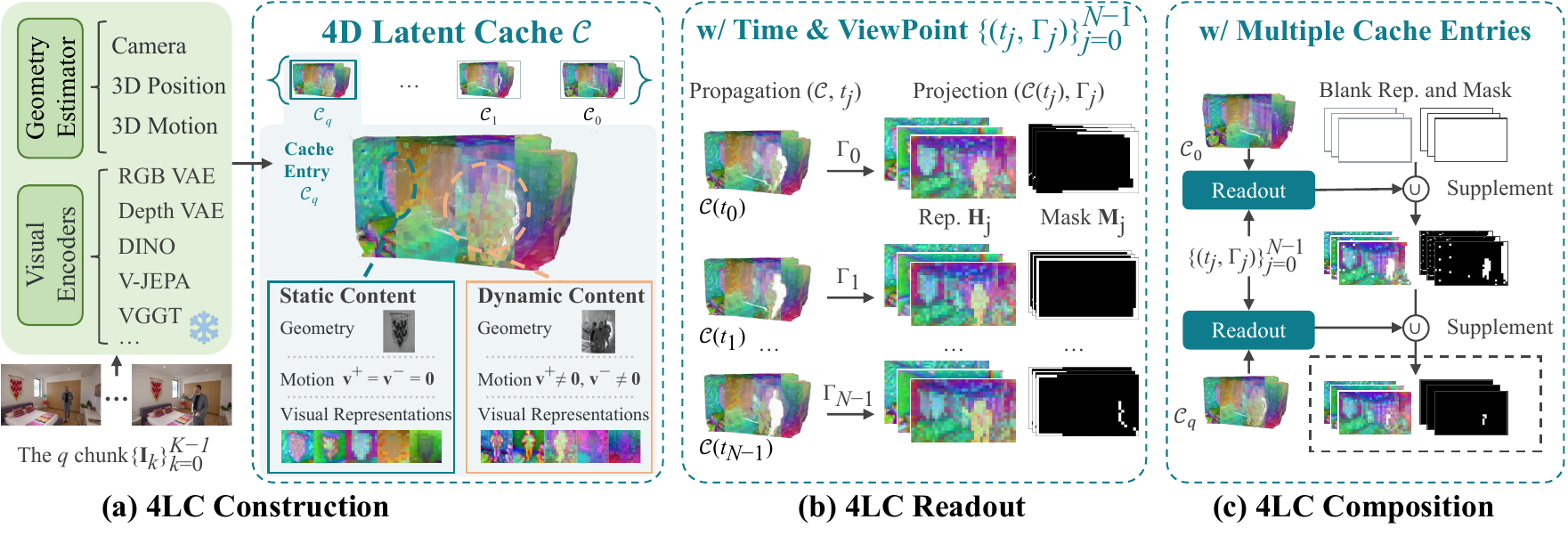}
    \par
    \caption{\textbf{Architecture of 4LC.} \textbf{(a) Construction} binds visual representations from multiple encoders to shared geometry and motion in one cache entry. \textbf{(b) Readout} propagates the entry to a requested time and projects it through a target camera, yielding readout features and coverage masks. \textbf{(c) Composition} fills uncovered regions with readouts from further entries.}
    \label{fig:generation-framework}
\end{figure}

The core of 4LC is a cache $\mathcal C=\{\mathcal C_q\}_{q=0}^{Q-1}$ and a spatiotemporal interface equipped with three operations.
Each entry $\mathcal C_q$ stores the visual representations of its source observations, bound to their 3D geometry and motion, and $\mathcal C$ organizes these entries by insertion order $q$.
\emph{Construction} encodes source observations with multiple visual encoders and stores their visual representations in a new cache entry (\autoref{fig:generation-framework}a).
\emph{Readout} propagates an entry's geometry to requested times and projects it through independently specified target cameras, reusing stored representations to produce readout features and coverage masks for static and dynamic content (\autoref{fig:generation-framework}b).
\emph{Composition} uses these masks to combine readouts across selected entries, filling uncovered regions while retaining existing readouts (\autoref{fig:generation-framework}c).
4LC-Video conditions a pretrained video diffusion model on these readouts (\autoref{sec:4lc-video}).
Appendix Sections~\ref{app:implementation-details} and~\ref{app:4lc-video-details} detail the implementation of 4LC and 4LC-Video, respectively.

\subsection{4LC Construction}
\label{sec:4lc-construction}

Construction builds an entry $\mathcal C_q=\{u_i\}_{i=0}^{P-1}$ of $P$ cache units from $K$ timestamped source images (keyframes) $\{\mathbf I_k\}_{k=0}^{K-1}$ in shared world coordinates and appends it to $\mathcal C$.
Each unit is defined as
\[
    u_i=\bigl(\mathbf x_i,t_i,\mathbf v_i^+,\mathbf v_i^-,b_i,
    \{\mathbf f_i^{(r)}\}_r\bigr).
\]
\paragraph{Geometry.}
Using depth and cameras from a pretrained estimator, we back-project source pixels to reference positions $\mathbf x_i\in\mathbb R^3$ in shared world coordinates.
Reference time $t_i$ is the source timestamp.

\paragraph{Motion.}
Motion estimates from the same estimator toward adjacent observations give local forward and backward world velocities $\mathbf v_i^+,\mathbf v_i^-\in\mathbb R^3$.
The motion mask $b_i\in\{0,1\}$ retains the estimated motion when $b_i=1$ and holds the reference position when $b_i=0$.
A single source image ($K=1$) provides no motion estimates between observations, so we set $b_i=0$.

\paragraph{Visual representations.}
Visual encoders produce visual representations as feature maps at different spatial and temporal resolutions.
For each visual representation $r$, we associate source frame $k$ with step $\alpha_r(k)$ of its encoded feature sequence $\mathbf Z^{(r)}$.
For unit $u_i$, its source frame $k_i\in\{0,\ldots,K-1\}$ and source-pixel coordinates $\mathbf p_i\in\mathbb R^2$ locate the bound feature vector:
\begin{equation}
    \mathbf F_k^{(r)}=\mathbf Z_{\alpha_r(k)}^{(r)},\qquad
    \mathbf f_i^{(r)}
    =\mathbf F_{k_i}^{(r)}
    \left[\left\lfloor\frac{\mathbf p_i}{d_r}\right\rfloor\right]
    \in\mathbb R^{D_r}.
    \label{eq:attribute-binding}
\end{equation}
Here, $d_r$ is the spatial stride and $D_r$ the feature dimension.
Units sharing a source feature cell reuse the same vector while retaining their own geometry and motion.

\subsection{4LC Readout}
\label{sec:4lc-transformation}

The target sequence $\{(t_j,\Gamma_j)\}_{j=0}^{N-1}$ specifies $N$ output frames, with scene time $t_j$ and camera configuration $\Gamma_j$ (intrinsics and world pose) at index $j$.
For a cache entry $\mathcal C_q$ and a pair $(t,\Gamma)$, the readout features and coverage for representation $r$ are
\begin{equation}
    \bigl(\mathbf H^{(r)},\mathbf M^{(r)}\bigr)
    =\Project\bigl(\Propagate(\mathcal C_q,t),\Gamma;r\bigr).
    \label{eq:spatiotemporal-readout}
\end{equation}

\paragraph{Temporal propagation.}
The local support $\mathcal T_i$ contains $t_i$ and the open intervals to adjacent source times with available motion estimates.
Under a locally linear approximation, the target position is
\begin{equation}
    \mathbf x_i(t)=
    \begin{cases}
        \mathbf x_i,
        & b_i=0\ \text{or}\ t=t_i,\\[2pt]
        \mathbf x_i+(t-t_i)\mathbf v_i^+,
        & b_i=1,\ t\in\mathcal T_i,\ t>t_i,\\[2pt]
        \mathbf x_i+(t_i-t)\mathbf v_i^-,
        & b_i=1,\ t\in\mathcal T_i,\ t<t_i.
    \end{cases}
    \label{eq:static-dynamic-position}
\end{equation}
The backward velocity points toward the past, hence the positive interval $t_i-t$.

The propagated entry $\mathcal C_q(t)=\Propagate(\mathcal C_q,t)$ retains units with defined target positions and preserves their feature bindings.
The source entry $\mathcal C_q$ remains unchanged and available for subsequent requests.
For $b_i=0$, the reference position remains defined beyond $\mathcal T_i$.
To suppress spurious motion on static structure, we set $b_i$ by thresholding projected displacement in the fixed source camera.
The threshold $\epsilon$ is measured in source-image pixels per video frame.

\paragraph{Camera projection.}
For each visual representation $r$, we project $\mathcal C_q(t)$ through $\Gamma$ onto its target grid, obtaining candidate units at each cell $\mathbf g\in\mathbb Z^2$.
Depth alone would let closer units from other times hide moving content, so we take moving candidates from the nearest source time first and fill remaining cells with static ones.
The selected unit index $i^\star(\mathbf g)$ determines the readout features:
\begin{equation}
    \mathbf H^{(r)}[\mathbf g]=\mathbf f_{i^\star(\mathbf g)}^{(r)}.
    \label{eq:indexed-readout}
\end{equation}
Coverage $\mathbf M^{(r)}[\mathbf g]=1$ when a unit is selected; otherwise, both are zero, and the coverage mask distinguishes missing content from valid zero-valued features.

An entry constructed from $K$ keyframes can provide conditions for an $N$-frame target sequence, including $N>K$, by reading each representation $r$ at its sampled target indices $\mathcal J_r\subseteq\{0,\ldots,N-1\}$ to obtain readouts $(\mathbf H_j^{(r)},\mathbf M_j^{(r)})$.
Different visual representations therefore share one access procedure with their own feature grids, dimensions, and temporal sampling.

\subsection{4LC Composition}
\label{sec:4lc-composition}

Composition applies Readout to selected cache entries in the specified order and combines their readouts to supplement coverage for the same target time and camera sequences (\autoref{fig:generation-framework}c).
For an entry $\mathcal C_q$, let $(\mathbf H_{q,j}^{(r)},\mathbf M_{q,j}^{(r)})$ denote its readout at target index $j$ for representation $r$.
The subscript $q$ retains the entry's insertion index in $\mathcal C$, independently of the order of access.
For each representation $r$ and $j\in\mathcal J_r$, we initialize the accumulated outputs as $\mathbf H_j^{(r)}=\mathbf0$ and $\mathbf M_j^{(r)}=\mathbf0$ on that representation's grid.
Each selected entry contributes through the updates
\begin{equation}
    \mathbf H_j^{(r)}\leftarrow\mathbf H_j^{(r)}
    +(\mathbf1-\mathbf M_j^{(r)})\odot\mathbf H_{q,j}^{(r)},\qquad
    \mathbf M_j^{(r)}\leftarrow\mathbf M_j^{(r)}\lor\mathbf M_{q,j}^{(r)}.
    \label{eq:composition-update}
\end{equation}
We apply the feature update before the coverage update, using the previous coverage to retain the first valid readout at overlapping locations.
The mask is broadcast across channels in the elementwise product $\odot$; $\lor$ denotes elementwise logical OR.
For each representation, the coverage ratio is the fraction of covered grid cells across its sampled target sequence.
Composition returns when every representation reaches the prescribed coverage-ratio threshold or no selected entries remain.
Entries therefore need not be accumulated into a single entry, avoiding the cost and redundancy of accumulation, and can be combined or evicted independently across requests.

\subsection{4LC-Video}
\label{sec:4lc-video}
\begin{figure}[tbp]
    \makeatletter
    \setlength{\@figindent}{0pt}
    \makeatother
    \centering
    \begin{minipage}[t]{0.42\linewidth}
        \vspace{0pt}
        \centering
        \captionof{table}[Task configurations for 4LC-Video.]{\textbf{Task configurations for 4LC-Video.} Requests use one entry $\mathcal C_q$ or selected entries from $\mathcal C$. $\bar t_j$ preserves the source or default single-image timeline, and $\bar\Gamma_j$ the source camera trajectory. Scene revisits use one readout along a camera path that returns to its start.}
        \label{tab:task-configuration}
        \begingroup
\small
\newcommand{\taskcell}[1]{\parbox[c][34.3pt][c]{\linewidth}{\raggedright #1}}
\begin{tabularx}{\linewidth}{@{}>{\raggedright\arraybackslash}X c c c@{}}
    \toprule
    Task
        & \begin{tabular}[c]{@{}c@{}}Initial\\frames\end{tabular}
        & \begin{tabular}[c]{@{}c@{}}Target\\request\\$(t_j,\Gamma_j)$\end{tabular}
        & \begin{tabular}[c]{@{}c@{}}Condition\\source\end{tabular} \\
    \midrule
    \taskcell{Camera control} & $\{\mathbf I_k\}_{k=0}^{K-1}$ & $(\bar t_j,\Gamma_j)$ & $\mathcal C_q$ \\
    \taskcell{Temporal control} & $\{\mathbf I_k\}_{k=0}^{K-1}$ & $(t_j,\bar \Gamma_j)$ & $\mathcal C_q$ \\
    \taskcell{Novel-view synthesis} & $\mathbf I_0$ & $(\bar t_j,\Gamma_j)$ & $\mathcal C_q$ \\
    \taskcell{Scene revisits} & $\mathbf I_0$ & $(\bar t_j,\Gamma_j)$ & $\mathcal C_q$ \\
    \taskcell{Long-term generation} & $\mathbf I_0$ & $(\bar t_j,\Gamma_j)$ & $\mathcal C$ \\
    \bottomrule
\end{tabularx}
\par\endgroup

    \end{minipage}\hfill
    \begin{minipage}[t]{0.56\linewidth}
        \vspace{0pt}
        \centering
        \vspace{\abovecaptionskip}
        \includegraphics[width=\linewidth]{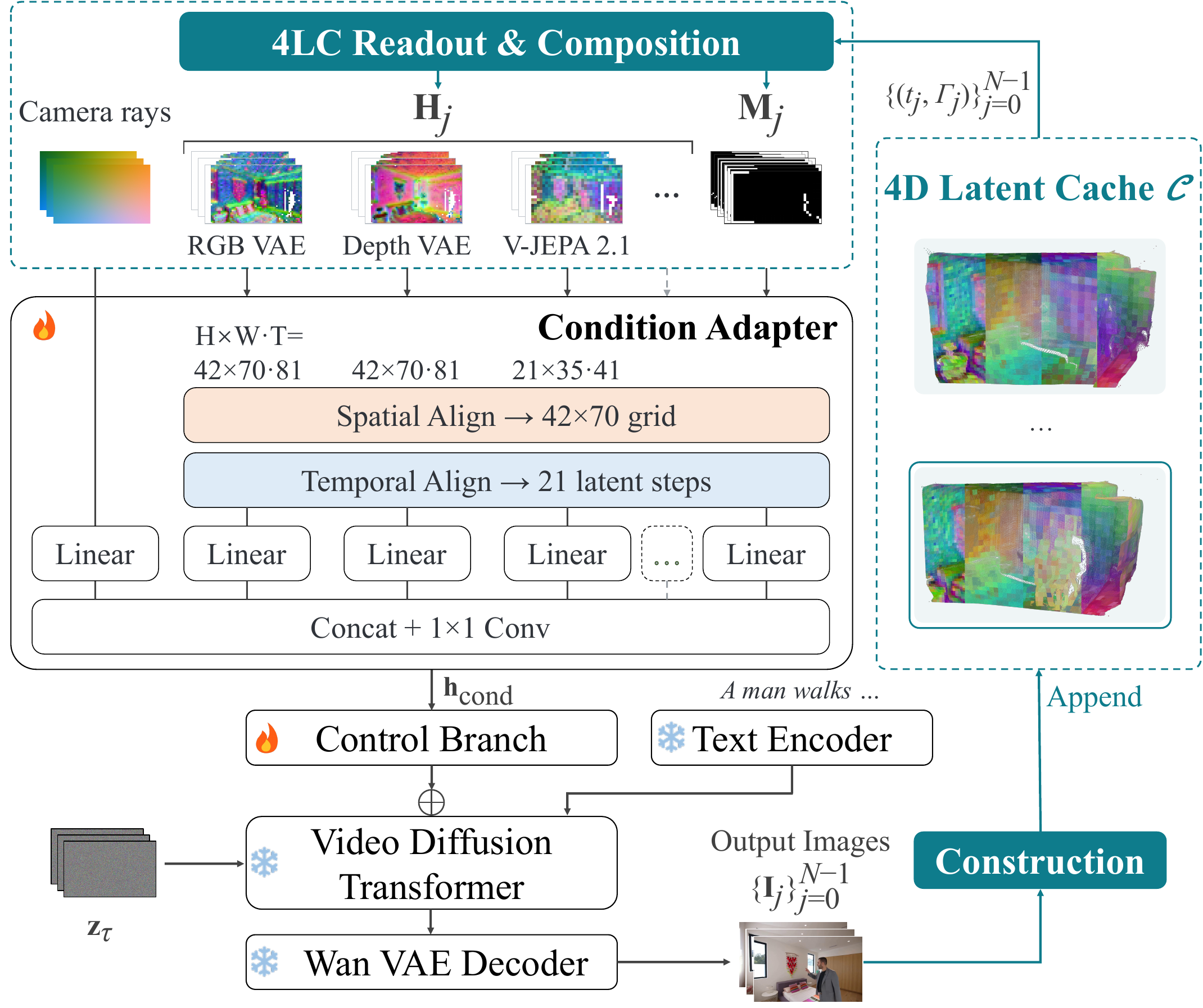}
        \captionof{figure}[4LC-Video framework.]{\textbf{4LC-Video framework.} The condition adapter aligns readouts to the Wan2.1 VAE latent grid and fuses them with masks and camera rays for the control branch. Generated observations extend the cache.}
        \label{fig:4lc-video}
    \end{minipage}
\end{figure}

\paragraph{Representation adaptation.}
Visual representations differ in spatial resolution, temporal sampling, and feature dimension.
A condition adapter aligns the outputs of Readout or Composition to the generator's spatiotemporal latent grid, using coverage masks to exclude missing values (\autoref{fig:4lc-video}).
Spatial alignment interpolates masked features and normalizes by interpolated coverage, keeping uncovered locations at zero.
Temporal alignment maps the initial readout to the first latent step and aggregates subsequent readouts within each latent time interval, averaging valid contributions and recording their coverage.
Separate linear projections map the aligned representations, merged coverage mask, and target-camera rays before concatenation and fusion into $\mathbf h_{\mathrm{cond}}$.

\paragraph{Conditional generation.}
A control branch injects this condition into the diffusion model to guide generation of target observations and complete uncovered regions.
Construction converts generated observations and their prescribed cameras and timestamps into new entries in $\mathcal C$ for reuse.
Because Readout takes times and cameras independently and Composition combines entries, the tasks in \autoref{tab:task-configuration} share one model and differ only in initial frames, target requests, and condition sources.

\paragraph{Training.}
\label{sec:training-augmentation}
We train the condition adapter and control branch with a weighted flow-matching objective~\citep{bib:flow_matching}, keeping the visual encoders and video diffusion backbone frozen.
Monocular training videos supply target sequences and sampled source observations, whose cache entries are queried at the original frame times and cameras.

\section{Experiments}
\label{sec:experiments}

\subsection{Experimental Setup}
\label{sec:experimental-setup}

\paragraph{Implementation details.}
4LC-Video uses Wan2.1-T2V-14B~\citep{bib:wan} with a VACE control branch~\citep{bib:vace} and the pretrained NeoVerse reconstructor~\citep{bib:neoverse} for geometry and motion estimation.
We select FLUX.2 VAE~\citep{bib:flux2} and V-JEPA 2.1~\citep{bib:vjepa21} as the default encoders for extracting visual representations.
Training uses 200K monocular clips from SpatialVID~\citep{bib:spatialvid}, RealCam-Vid~\citep{bib:realcam_vid}, and OpenVid~\citep{bib:openvid} without paired multiview supervision.
Inference generates 81-frame $560\times336$ segments with the LightX2V 4-step LoRA~\citep{bib:lightx2v}.

\paragraph{Benchmarks and baselines.}
Camera control compares with SpaceTimePilot~\citep{bib:spacetimepilot}, TrajectoryCrafter~\citep{bib:trajectorycrafter}, GEN3C~\citep{bib:gen3c}, and NeoVerse~\citep{bib:neoverse}, and temporal control with SpaceTimePilot and NeoVerse.
Both tasks report camera errors and six VBench~\citep{bib:vbench} scores.
Novel-view synthesis on RealEstate10K~\citep{bib:realestate10k} and scene revisits on WorldScore~\citep{bib:worldscore} compare with ViewCrafter~\citep{bib:viewcrafter}, FlexWorld~\citep{bib:flexworld}, Voyager~\citep{bib:voyager}, Spatia~\citep{bib:spatia}, Mirage~\citep{bib:mirage}, and LiveWorld~\citep{bib:liveworld} with PSNR, SSIM~\citep{bib:ssim}, and LPIPS~\citep{bib:lpips}.
Long-term generation on MBench-A compares with published results~\citep{bib:mbench} of HY-WorldPlay~\citep{bib:hy_worldplay}, Infinite-World~\citep{bib:infinite_world}, Yume-1.5~\citep{bib:yume15}, and LingBot-World~\citep{bib:lingbot_world}.
\appref{app:experimental-details} and \appref{app:additional-results} provide evaluation protocols and additional results.

\begingroup
\makeatletter
\renewcommand{\subsection}{\@startsection{subsection}{2}{\z@}
  {9pt plus 2pt minus 1pt}{9pt plus 1pt minus 1pt}{\elvbf}}
\renewcommand{\paragraph}{\@startsection{paragraph}{4}{\z@}
  {6pt plus 2pt minus 1pt}{-1em}{\normalfont\normalsize\bfseries}}
\makeatother

\subsection{Main Results on Multiple Video Generation Tasks}
\label{sec:multi-task-generation}

We evaluate a single 4LC-Video checkpoint on the five tasks in \autoref{tab:task-configuration} without task-specific fine-tuning.
Camera and temporal control read 4LC built from a source video with static and dynamic content, novel-view synthesis and scene revisits read an entry built from a single image, and long-term generation reads entries accumulated across generated segments.

\begin{figure}[tbp]
    \centering
    \includegraphics[width=\linewidth]{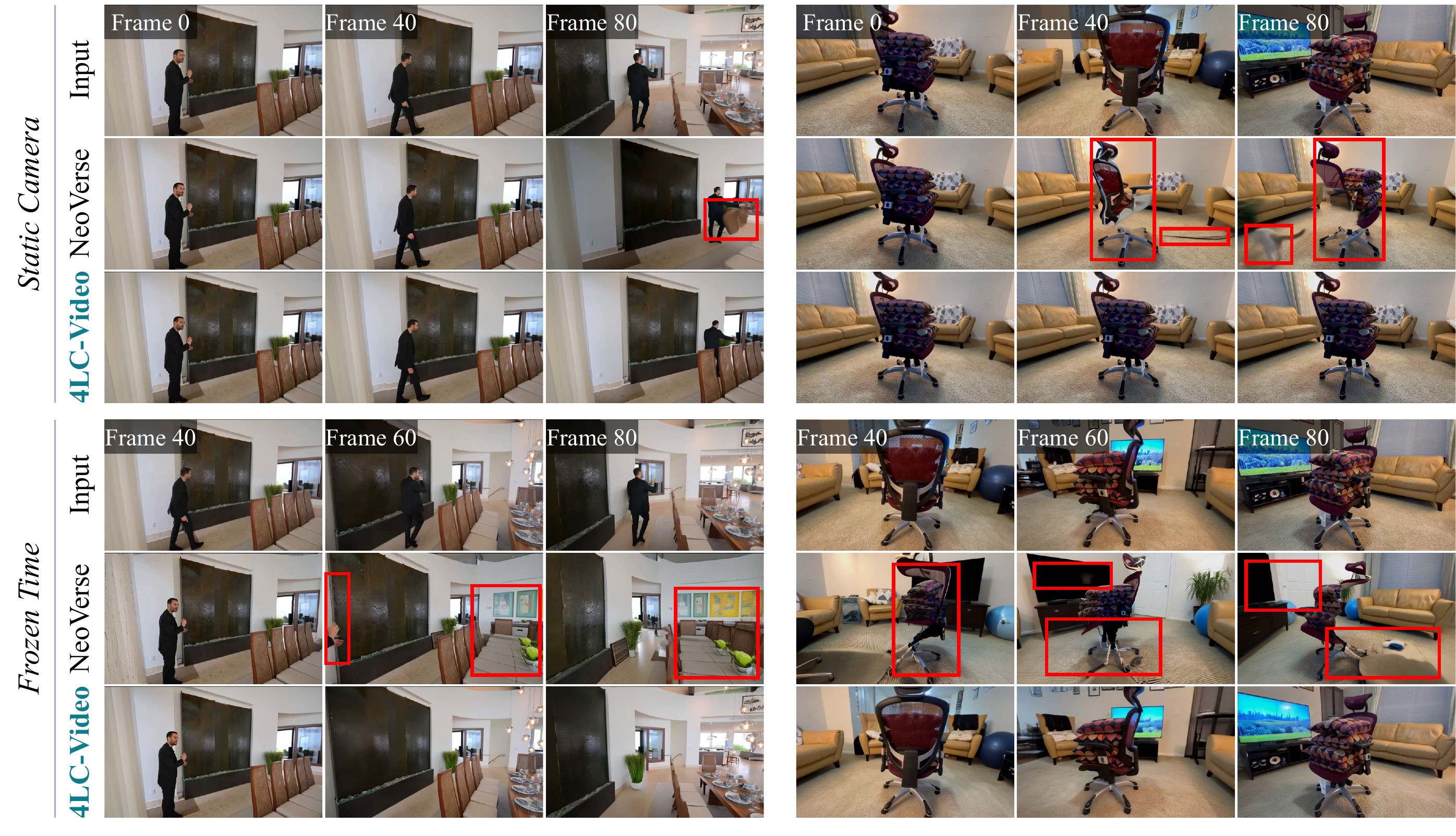}
    \par
    \caption{\textbf{Qualitative comparison for camera and temporal control.} Left: dynamic scene; right: static scene. \textbf{Top: Static camera.} The camera is fixed while scene time follows the source timeline. \textbf{\mbox{Bottom:~Frozen} time.} Scene time is fixed at source frame 0 while the camera follows the source trajectory. Input frame labels identify each column; red boxes highlight distortions.}
    \label{fig:qualitative-control}
\end{figure}

\paragraph{Camera control.}
4LC-Video attains the lowest camera errors together with the highest subject and background consistency (\autoref{tab:spatiotemporal-control}), whereas the baselines trade one for the other: NeoVerse follows the cameras with lower consistency, GEN3C and SpaceTimePilot stay consistent but close to static views with several-fold camera errors, and TrajectoryCrafter trails on both.
4LC-Video avoids this trade-off because 4LC keeps static structure persistent and dynamic content accurate.
GEN3C, TrajectoryCrafter, and NeoVerse instead render conditions from reconstructed point clouds or 4DGS, whose artifacts and holes appear on moving subjects and static structures in NeoVerse outputs, while 4LC-Video stays faithful to the source observations (\autoref{fig:qualitative-control}).

\paragraph{Temporal control.}
4LC-Video keeps this advantage under temporal control (\autoref{tab:spatiotemporal-control}), while the baselines show the same trade-off.
For time and camera pairs never observed together, 4LC reads each moving subject from its own observations and moves it by its estimated motion, while static structure stays in place.
SpaceTimePilot generates the retimed motion without explicit conditions, and NeoVerse renders it from 4DGS with the artifacts noted above.
In the frozen-time examples of \autoref{fig:qualitative-control}, 4LC-Video holds the subject at frame 0 and keeps later-observed background intact, whereas NeoVerse deforms the subject inconsistently.

\begin{table}[tbp]
    \centering
    \caption{\textbf{Quantitative comparison for camera and temporal control.} We evaluate 100 randomly sampled unseen videos under four camera trajectories (400 cases) and three temporal trajectories (300 cases). $r_{\mathrm{dist}}$ measures rotation error in radians and $t_{\mathrm{dist}}$ scale-normalized translation error; higher VBench scores are better. Best and second-best results per task are bold and underlined.}
    \label{tab:spatiotemporal-control}

    \small
    \begin{tabular}{@{}lcccccccc@{}}
        \toprule
        Method & $r_{\mathrm{dist}}{\downarrow}$ & $t_{\mathrm{dist}}{\downarrow}$ & \begin{tabular}[c]{@{}c@{}}Subj.\\Consist.\end{tabular} & \begin{tabular}[c]{@{}c@{}}Back.\\Consist.\end{tabular} & \begin{tabular}[c]{@{}c@{}}Temp.\\Flick.\end{tabular} & \begin{tabular}[c]{@{}c@{}}Motion\\Smooth.\end{tabular} & \begin{tabular}[c]{@{}c@{}}Aesth.\\Quality\end{tabular} & \begin{tabular}[c]{@{}c@{}}Imag.\\Quality\end{tabular} \\
        \midrule
        \multicolumn{9}{l}{\textit{Camera control}} \\
        SpaceTimePilot & 0.179 & 0.321 & 89.79 & 93.43 & 95.26 & \textbf{98.69} & 47.60 & 57.23 \\
        TrajectoryCrafter & 0.244 & 0.395 & 86.37 & 92.19 & 92.69 & 96.17 & 45.16 & 60.97 \\
        GEN3C & 0.203 & 0.285 & \underline{90.45} & \underline{93.79} & \textbf{95.66} & 97.55 & 45.62 & 56.39 \\
        NeoVerse & \underline{0.059} & \underline{0.120} & 88.63 & 92.77 & 94.43 & \underline{98.17} & \textbf{50.38} & \textbf{67.84} \\
        \midrule
        \rowcolor[HTML]{D9E1F2} 4LC-Video (ours) & \textbf{0.037} & \textbf{0.089} & \textbf{90.71} & \textbf{94.14} & \underline{95.47} & 98.06 & \underline{48.16} & \underline{65.49} \\
        \midrule
        \multicolumn{9}{l}{\textit{Temporal control}} \\
        SpaceTimePilot & 0.163 & 0.192 & \underline{88.42} & \underline{92.65} & \textbf{94.64} & \textbf{98.37} & 47.44 & 56.40 \\
        NeoVerse & \underline{0.057} & \underline{0.080} & 87.55 & 92.11 & 93.38 & \underline{97.88} & \textbf{49.65} & \textbf{66.85} \\
        \midrule
        \rowcolor[HTML]{D9E1F2} 4LC-Video (ours) & \textbf{0.041} & \textbf{0.060} & \textbf{88.58} & \textbf{92.85} & \underline{93.78} & 97.41 & \underline{48.75} & \underline{65.14} \\
        \bottomrule
    \end{tabular}
\end{table}

\Needspace{18\baselineskip}
\begingroup
\setlength{\columnsep}{14pt}
\begin{wraptable}{r}{0.56\textwidth}
    \vspace{-\dimexpr\intextsep+\abovecaptionskip\relax}
    \centering
    \caption{\textbf{Quantitative comparison for novel-view synthesis and scene revisits.} We use the protocols and baseline results reported by Spatia~\citep{bib:spatia} and Mirage~\citep{bib:mirage}, and evaluate LiveWorld under the same protocols.}
    \label{tab:memory-quality}

    \small
    \setlength{\tabcolsep}{2.5pt}
    \begin{tabular}{@{}lcccccc@{}}
        \toprule
        \multirow{2}{*}{Method} & \multicolumn{3}{c}{Novel-view synthesis} & \multicolumn{3}{c}{Closed-loop consistency} \\
        \cmidrule(lr){2-4}\cmidrule(l){5-7}
         & PSNR${\uparrow}$ & SSIM${\uparrow}$ & LPIPS${\downarrow}$ & PSNR$_C{\uparrow}$ & SSIM$_C{\uparrow}$ & LPIPS$_C{\downarrow}$ \\
        \midrule
        ViewCrafter & 15.78 & 0.580 & 0.396 & 14.79 & 0.481 & 0.365 \\
        FlexWorld & 16.25 & 0.593 & 0.370 & 12.20 & 0.428 & 0.598 \\
        Voyager & 17.79 & 0.636 & 0.297 & 17.66 & 0.540 & 0.380 \\
        Spatia & 18.58 & 0.646 & 0.254 & 19.38 & 0.579 & 0.213 \\
        Mirage & 18.38 & \textbf{0.779} & 0.250 & 20.05 & \textbf{0.825} & 0.228 \\
        LiveWorld & \textbf{19.04} & 0.645 & \underline{0.191} & \underline{20.07} & 0.653 & \underline{0.186} \\
        \midrule
        \rowcolor[HTML]{D9E1F2} 4LC-Video (ours) & \underline{18.94} & \underline{0.650} & \textbf{0.188} & \textbf{21.91} & \underline{0.705} & \textbf{0.161} \\
        \bottomrule
    \end{tabular}
\end{wraptable}

\paragraph{Novel-view synthesis and \mbox{Scene revisits.}}
4LC-Video obtains the best novel-view LPIPS with PSNR within 0.1 dB of LiveWorld, and it exceeds ViewCrafter, FlexWorld, Voyager, and Spatia on all three metrics (\autoref{tab:memory-quality}).
For static scenes, 4LC reuses the cached visual representations of the image without RGB rendering and re-encoding, so its conditions avoid the information loss of these steps.
Mirage attains higher SSIM at a higher output resolution ($1280\times704$ versus $560\times336$), while its PSNR and LPIPS are worse than those of 4LC-Video.
Scene revisits expose accumulated drift.
4LC-Video leads LiveWorld by 1.84 dB in PSNR$_C$ and attains the best LPIPS$_C$.
The baselines update their point clouds or spatial memories with generated views along the path, allowing intermediate errors to re-enter the conditions of return views.
Reusing one source-image entry, 4LC gives each revisited camera the same condition regardless of intermediate outputs.
\par\endgroup

\suppressfloats[t]
\begin{table}[tbp]
    \centering
    \caption{\textbf{Quantitative comparison for long-term generation on MBench-A~\citep{bib:mbench}.} We compare Environment and Causal scores with published leaderboard results.}
    \label{tab:long-horizon-memory}

    \small
    \begin{tabular}{@{}lccccccccc@{}}
        \toprule
        \multirow{2}{*}{Method} & \multirow{2}{*}{\shortstack{Env.\\score}} & \multirow{2}{*}{\shortstack{Causal\\score}} & \multicolumn{2}{c}{Spatial} & \multicolumn{2}{c}{Rendering} & \multicolumn{2}{c}{Self-evolution} & Interaction \\
        \cmidrule(lr){4-5}\cmidrule(lr){6-7}\cmidrule(lr){8-9}\cmidrule(l){10-10}
         & & & Epipolar & Reproj. & Lighting & Style & State & Correct. & Action \\
        \midrule
        HY-WorldPlay & \underline{83.23} & 65.86 & \textbf{83.86} & \underline{68.17} & \underline{82.67} & \textbf{98.23} & 49.50 & 62.40 & 85.69 \\
        Infinite-World & 73.76 & 70.92 & \underline{74.04} & 61.51 & 62.63 & 96.87 & 48.00 & 78.40 & \underline{86.37} \\
        Yume-1.5 & 54.92 & \underline{85.03} & 51.86 & 24.55 & 51.21 & 92.05 & \textbf{97.90} & \textbf{95.00} & 62.20 \\
        LingBot-World & 38.90 & 82.91 & 22.12 & 7.57 & 40.06 & 85.87 & \underline{96.00} & 89.40 & 63.32 \\
        \midrule
        \rowcolor[HTML]{D9E1F2} 4LC-Video (ours) & \textbf{85.56} & \textbf{89.06} & 71.11 & \textbf{75.98} & \textbf{97.52} & \underline{97.66} & 85.00 & \underline{93.20} & \textbf{88.97} \\
        \bottomrule
    \end{tabular}
\end{table}

\paragraph{Long-term generation.}
Reuse across generated segments separates the MBench-A baselines, which either preserve the environment or evolve the scene (\autoref{tab:long-horizon-memory}).
HY-WorldPlay and Infinite-World score high on spatial and rendering metrics but low on state evolution, as their scenes remain overly static~\citep{bib:mbench}.
Yume-1.5 and LingBot-World evolve scene states but lose spatial consistency.
4LC-Video bridges the two with the highest Environment and Causal scores.
Each generated segment is appended to 4LC as a new entry, and later segments reuse only the static units of earlier entries as history through Composition.
Revisited structure is thus constrained, while past motion is not replayed and scene states can keep evolving.
Return views reproduce the outbound content, whereas HY-WorldPlay shifts the layout of revisited regions (\autoref{fig:app-mbench} in \appref{app:long-term-qualitative}).

\subsection{Conditioning Efficiency and Quality}
\label{sec:readouts-cost}

\begin{figure}[tp]
    \centering
    \includegraphics[width=\linewidth]{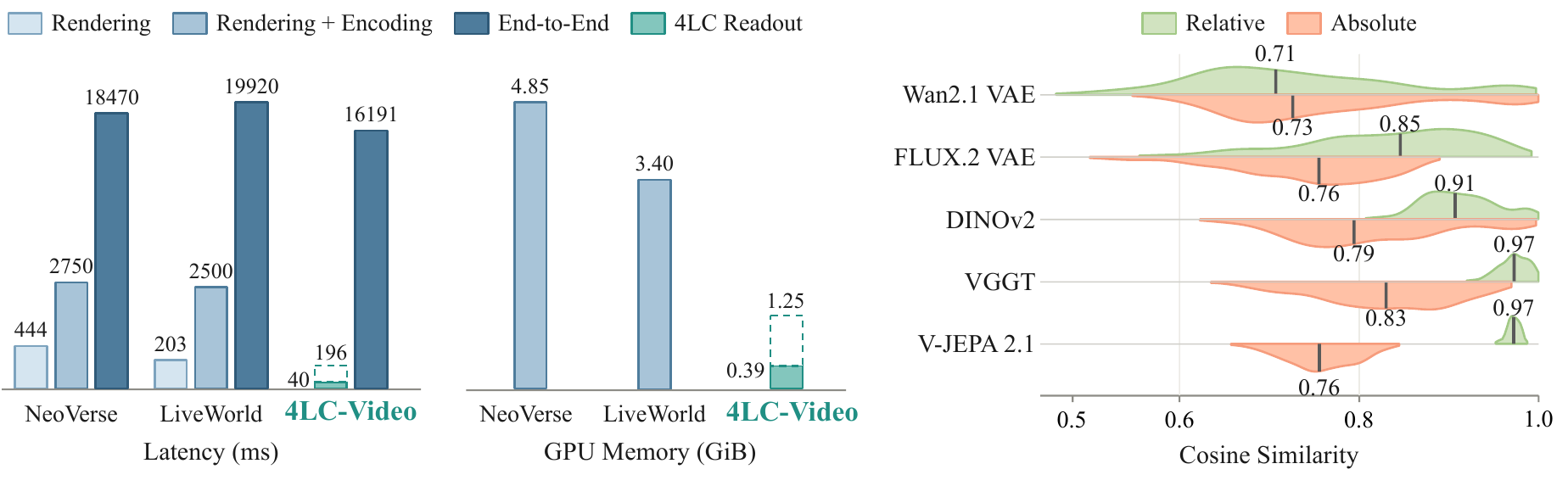}
    \par
    \caption{\textbf{Efficiency and feature similarity across condition-acquisition paths.} \textbf{Left: Latency and peak GPU memory} of 4LC readout and of RGB rendering and VAE encoding from NeoVerse's 4DGS and LiveWorld's RGB point clouds, with the end-to-end latency of each framework. Rendering + Encoding includes rendering time. Solid and dashed 4LC Readout bars use Wan2.1 VAE alone and with DINOv2, VGGT, and V-JEPA 2.1. Latency uses a square-root scale. \textbf{Right: Feature similarity} between direct readouts and re-encoded features, as per-case mean cosine similarity in the original (Absolute) and relative-representation (Relative) spaces; black ticks mark medians.}
    \label{fig:readouts-cost}
\end{figure}

\paragraph{Conditioning efficiency.}
Condition acquisition accounts for 14.9\% of the 18.47 s end-to-end generation time in NeoVerse, compared with 0.25\% of 16.19 s for 4LC-Video under the same generation settings, and this overhead recurs across segments in long-term generation.
In \autoref{fig:readouts-cost}, NeoVerse, LiveWorld, and 4LC-Video all use the Wan2.1 backbone and VAE encoder for 81-frame requests at $560\times336$ (\appref{app:efficiency-protocol}).
The conditioning paths in NeoVerse~\citep{bib:neoverse} and LiveWorld~\citep{bib:liveworld} render image-resolution RGB views from 4DGS and RGB point clouds, respectively, followed by VAE encoding (\autoref{fig:readouts-cost}, left).
4LC readout directly projects cached representations onto lower-resolution feature grids without repeated encoding.
Compared with 4DGS rendering followed by VAE encoding in NeoVerse, 4LC readout is $68.5\times$ faster, with $12.6\times$ lower peak GPU memory.
Adding DINOv2, VGGT, and V-JEPA 2.1 to Wan2.1 VAE raises readout latency from 40.14 to 196.14 ms and peak GPU memory from 0.385 to 1.245 GiB.
The default FLUX.2 VAE with V-JEPA 2.1 takes 162.03~ms and 0.763~GiB, and adding DINOv2 and VGGT raises them to 255.20~ms and 1.605~GiB.
All remain below the cost of rendering and encoding one representation.

\paragraph{Conditioning quality.}
To test whether direct readout preserves the information of re-encoding, we compare it with re-encoding views rendered from cached source pixels under the same Readout and encoders.
All features retain comparable absolute similarity (\autoref{fig:readouts-cost}, right), while semantic and geometric features also preserve stable relative structure~\citep{bib:latent_analysis}, with VGGT~\citep{bib:vggt} and V-JEPA 2.1 concentrating near one in relative space and DINOv2~\citep{bib:dinov2} showing intermediate similarity.
VAE features show lower relative similarity (medians 0.85 for FLUX.2 VAE and 0.71 for Wan2.1 VAE), as projection, resampling, and re-encoding alter their pixel-level detail.
\autoref{sec:ablation-studies} examines the effect of condition-acquisition paths on generation.

\subsection{Ablation Studies}
\label{sec:ablation-studies}
\begingroup
\setlength{\columnsep}{14pt}
\begin{wraptable}{r}{0.53\textwidth}
    \vspace{-\dimexpr\intextsep+\abovecaptionskip\relax}
    \centering
    \caption{\textbf{Ablations of condition-acquisition paths and composite visual representations.} All variants share a frozen Wan2.1-T2V-1.3B backbone and the same 100 unseen videos with all trajectories. ALL denotes the other three encoders.}
    \label{tab:representation-combinations}

    \small
    \begin{tabular}{@{}lcccc@{}}
        \toprule
        Visual representations & $r_{\mathrm{dist}}{\downarrow}$ & $t_{\mathrm{dist}}{\downarrow}$ & \begin{tabular}[c]{@{}c@{}}Subj.\\Consist.\end{tabular} & \begin{tabular}[c]{@{}c@{}}Back.\\Consist.\end{tabular} \\
        \midrule
        \multicolumn{5}{l}{\textit{RGB rendering and re-encoding}} \\
        Wan2.1 VAE & \underline{0.046} & \underline{0.106} & 89.86 & 93.69 \\
        FLUX.2 VAE & \textbf{0.044} & \underline{0.106} & 90.00 & 93.80 \\
        \midrule
        \multicolumn{5}{l}{\textit{Direct readout}} \\
        Wan2.1 VAE & 0.056 & 0.128 & 89.03 & 93.03 \\
        FLUX.2 VAE & 0.049 & 0.114 & \underline{90.06} & 93.81 \\
        \midrule
        \multicolumn{5}{l}{\textit{Direct readout with composite representations}} \\
        FLUX.2 VAE + DINOv2 & 0.051 & 0.116 & 89.89 & 93.69 \\
        FLUX.2 VAE + VGGT & 0.048 & 0.120 & 89.89 & 93.70 \\
        FLUX.2 VAE + V-JEPA 2.1 & \underline{0.046} & 0.112 & 90.05 & \underline{93.85} \\
        FLUX.2 VAE + ALL & \textbf{0.044} & \textbf{0.101} & \textbf{90.15} & \textbf{93.90} \\
        \bottomrule
    \end{tabular}
\end{wraptable}

\paragraph{Condition-acquisition path.}
For each VAE, we compare direct readout with RGB rendering and re-encoding (\autoref{tab:representation-combinations}).
Re-encoding updates features in the rendered target context, while direct readout retains the source encoding as content moves and visibility changes.
With FLUX.2 VAE, direct readout leaves subject and background consistency nearly unchanged, with a small difference in camera errors.
For Wan2.1 VAE, direct readout lowers both consistency scores and raises camera errors more.
This matches FLUX.2 VAE's higher readout similarity in \autoref{sec:readouts-cost} and supports its selection.

\paragraph{Composite visual representations.}
A single visual representation cannot meet the representational demands of dynamic world modeling, which span appearance, semantics, geometry, and dynamics, whereas existing encoders each specialize in one of them: FLUX.2 VAE reconstructs appearance, DINOv2 learns semantic features, VGGT infers multi-view geometry, and V-JEPA 2.1 models video dynamics.
Since 4LC binds these representations to shared geometry and motion, we add DINOv2, VGGT, and V-JEPA 2.1 to FLUX.2 VAE under direct readout, individually and together (\autoref{tab:representation-combinations}).
Individually, DINOv2 and VGGT keep camera errors and consistency close to FLUX.2 VAE, indicating that their semantic and geometric cues are not fully exploited in pairwise combinations.
V-JEPA 2.1 lowers both rotation and translation errors while leaving consistency nearly unchanged: unlike the image and multi-view encoders, it learns from video, and its temporally consistent features~\citep{bib:vjepa21} keep high similarity after readout, bringing temporal context into the condition.
Combining all four lets appearance, semantic, geometric, and dynamic cues complement one another, making it the strongest configuration: it reduces the translation error of FLUX.2 VAE combined with V-JEPA 2.1 from 0.112 to 0.101, and all four metrics match or exceed FLUX.2 VAE re-encoding.
Because Readout reuses cached representations, it adds no efficiency bottleneck (\autoref{fig:readouts-cost}, \autoref{sec:readouts-cost}), so 4LC is both composite and efficient, yielding rich, high-quality 4D conditions.
\par\endgroup
\endgroup

\begingroup
\makeatletter
\renewcommand{\section}{\@startsection{section}{1}{\z@}
  {10pt plus 2pt minus 1pt}{10pt plus 1pt minus 1pt}{\large\bfseries}}
\makeatother
\section{Conclusion}
We presented 4LC, a 4D latent cache that stores composite visual representations of static and dynamic content and reads them directly at independently requested times and viewpoints.
Conditioned on these readouts, a single 4LC-Video checkpoint achieved state-of-the-art control accuracy and scene consistency across five generation tasks.
By avoiding per-request RGB rendering and re-encoding, 4LC turns condition acquisition from a recurring cost into a one-time encoding of observations.
Its composite representations add complementary information from multiple encoders, giving dynamic world modeling conditions that are both efficient and rich.

\paragraph{Limitations and future work.}
4LC relies on a feed-forward 4D reconstruction model to estimate depth, camera parameters, and motion, and its estimation errors can affect readout quality despite training augmentation.
While 4LC supports both static and dynamic content, 4LC-Video reuses only static historical content for long-term generation.
We leave dynamic reuse across generation segments to future work, as a further step toward efficient conditioning for multi-task video generation.
\par\endgroup

\bibliographystyle{unsrtnat}
\bibliography{main}

\begin{thebibliography}{72}
\providecommand{\natexlab}[1]{#1}
\providecommand{\url}[1]{\texttt{#1}}
\expandafter\ifx\csname urlstyle\endcsname\relax
  \providecommand{\doi}[1]{doi: #1}\else
  \providecommand{\doi}{doi: \begingroup \urlstyle{rm}\Url}\fi

\bibitem[Yao et~al.(2025)Yao, Zhai, and Wang]{bib:uni4d}
David~Yifan Yao, Albert~J. Zhai, and Shenlong Wang.
\newblock {Uni4D}: Unifying visual foundation models for {4D} modeling from a
  single video.
\newblock In \emph{CVPR}, 2025.

\bibitem[Qiao et~al.(2026)Qiao, An, Xiong, Belongie, and
  Jacobs]{bib:track2view}
Feng Qiao, Zhaochong An, Zhexiao Xiong, Serge Belongie, and Nathan Jacobs.
\newblock {Track2View}: {4D-Consistent} camera-controlled video generation via
  paired {3D} point tracks.
\newblock \emph{arXiv preprint arXiv:2606.15534}, 2026.

\bibitem[Fang et~al.(2026)Fang, Jiang, Bai, Mitra, and Huang]{bib:worldreel}
Shaoheng Fang, Hanwen Jiang, Yunpeng Bai, Niloy~J. Mitra, and Qixing Huang.
\newblock {WorldReel}: {4D} video generation with consistent geometry and
  motion modeling.
\newblock In \emph{CVPR}, 2026.

\bibitem[Ren et~al.(2025)Ren, Shen, Huang, Ling, Lu, Nimier-David, M\"uller,
  Keller, Fidler, and Gao]{bib:gen3c}
Xuanchi Ren, Tianchang Shen, Jiahui Huang, Huan Ling, Yifan Lu, Merlin
  Nimier-David, Thomas M\"uller, Alexander Keller, Sanja Fidler, and Jun Gao.
\newblock {GEN3C}: {3D-Informed} world-consistent video generation with precise
  camera control.
\newblock In \emph{CVPR}, 2025.

\bibitem[Zhao et~al.(2026)Zhao, Wei, Liu, Zhang, Xu, and Lu]{bib:spatia}
Jinjing Zhao, Fangyun Wei, Zhening Liu, Hongyang Zhang, Chang Xu, and Yan Lu.
\newblock Spatia: Video generation with updatable spatial memory.
\newblock In \emph{CVPR}, 2026.

\bibitem[Wu et~al.(2025)Wu, Yang, Po, Xu, Liu, Lin, and
  Wetzstein]{bib:long_term_spatial_memory}
Tong Wu, Shuai Yang, Ryan Po, Yinghao Xu, Ziwei Liu, Dahua Lin, and Gordon
  Wetzstein.
\newblock Video world models with long-term spatial memory.
\newblock In \emph{NeurIPS}, 2025.

\bibitem[Duan et~al.(2026)Duan, Xia, Zhang, Zhang, Zhou, Gou, He, Chen, Zhang,
  and Liu]{bib:liveworld}
Zicheng Duan, Jiatong Xia, Zeyu Zhang, Wenbo Zhang, Gengze Zhou, Chenhui Gou,
  Yefei He, Feng Chen, Xinyu Zhang, and Lingqiao Liu.
\newblock {LiveWorld}: Simulating out-of-sight dynamics in generative video
  world models.
\newblock In \emph{ECCV}, 2026.

\bibitem[Zheng et~al.(2026)Zheng, Yin, Hu, Li, Shan, and Fu]{bib:versecrafter}
Sixiao Zheng, Minghao Yin, Wenbo Hu, Xiaoyu Li, Ying Shan, and Yanwei Fu.
\newblock {VerseCrafter}: Dynamic realistic video world model with {4D}
  geometric control.
\newblock In \emph{CVPR}, 2026.

\bibitem[Wu et~al.(2024)Wu, Yi, Fang, Xie, Zhang, Wei, Liu, Tian, and
  Wang]{bib:4dgs}
Guanjun Wu, Taoran Yi, Jiemin Fang, Lingxi Xie, Xiaopeng Zhang, Wei Wei, Wenyu
  Liu, Qi~Tian, and Xinggang Wang.
\newblock {4D} {G}aussian splatting for real-time dynamic scene rendering.
\newblock In \emph{CVPR}, 2024.

\bibitem[Yang et~al.(2026)Yang, Fan, Shi, Peng, Wang, and Zhang]{bib:neoverse}
Yuxue Yang, Lue Fan, Ziqi Shi, Junran Peng, Feng Wang, and Zhaoxiang Zhang.
\newblock {NeoVerse}: Enhancing {4D} world model with in-the-wild monocular
  videos.
\newblock In \emph{CVPR}, 2026.

\bibitem[Yi et~al.(2026)Yi, Kim, Cho, Jang, Yun, and Kim]{bib:worldkv}
Jung Yi, Minjae Kim, Paul~Hyunbin Cho, Wooseok Jang, Sangdoo Yun, and
  Seungryong Kim.
\newblock {WorldKV}: Efficient world memory with world retrieval and
  compression.
\newblock \emph{arXiv preprint arXiv:2605.22718}, 2026.

\bibitem[Feng et~al.(2026)Feng, Fan, Qin, Wu, Li, Li, An, Huang, Wang, Liao,
  Magno, Xu, and Yang]{bib:worldcache}
Weilun Feng, Guoxin Fan, Haotong Qin, Mingqiang Wu, Yuqi Li, Xiangqi Li, Zhulin
  An, Libo Huang, Dingrui Wang, Longlong Liao, Michele Magno, Yongjun Xu, and
  Chuanguang Yang.
\newblock {WorldCache}: Accelerating world models for free via heterogeneous
  token caching.
\newblock In \emph{ICML}, 2026.

\bibitem[Wang et~al.(2026{\natexlab{a}})Wang, Zhao, Yang, Chen, Zhang, He,
  Duan, Chen, Yang, and Zhuang]{bib:mirage}
Weijie Wang, Haoyu Zhao, Yifan Yang, Feng Chen, Zeyu Zhang, Yefei He, Zicheng
  Duan, Donny~Y. Chen, Yuqing Yang, and Bohan Zhuang.
\newblock Latent spatial memory for video world models.
\newblock \emph{arXiv preprint arXiv:2606.09828}, 2026{\natexlab{a}}.

\bibitem[Zhu et~al.(2025)Zhu, Wang, Zhou, Chang, Zhou, Li, Chen, Shen, Pang,
  and He]{bib:aether}
Haoyi Zhu, Yifan Wang, Jianjun Zhou, Wenzheng Chang, Yang Zhou, Zizun Li, Junyi
  Chen, Chunhua Shen, Jiangmiao Pang, and Tong He.
\newblock Aether: Geometric-aware unified world modeling.
\newblock In \emph{ICCV}, 2025.

\bibitem[Chen et~al.(2025{\natexlab{a}})Chen, Liang, Wang, Chen, Cheng, Gu,
  Huang, Jiang, Li, Li, Li, Li, Liu, Liu, Liu, Wang, Weng, Wu, Xiang, Yang,
  Zhang, Zhang, Zhou, Zhou, Huang, Zhang, and Li]{bib:teleworld}
Yabo Chen, Yuanzhi Liang, Jiepeng Wang, Tingxi Chen, Junfei Cheng, Zixiao Gu,
  Yuyang Huang, Zicheng Jiang, Wei Li, Tian Li, Weichen Li, Zuoxin Li, Guangce
  Liu, Jialun Liu, Junqi Liu, Haoyuan Wang, Qizhen Weng, Xuan'er Wu, Xunzhi
  Xiang, Xiaoyan Yang, Xin Zhang, Shiwen Zhang, Junyu Zhou, Chengcheng Zhou,
  Haibin Huang, Chi Zhang, and Xuelong Li.
\newblock {TeleWorld}: Towards dynamic multimodal synthesis with a {4D} world
  model.
\newblock \emph{arXiv preprint arXiv:2601.00051}, 2025{\natexlab{a}}.

\bibitem[{InSpatio Team} et~al.(2026){InSpatio Team}, Shen, Zhang, Liu, Ji,
  Bao, Zhai, Liu, Guo, Wang, Pan, Pan, Xie, Liu, Xiang, Zhang, Chen, Wang,
  Chen, Fan, Le, Ye, and Zhao]{bib:inspatio_world}
{InSpatio Team}, Donghui Shen, Guofeng Zhang, Haomin Liu, Haoyu Ji, Hujun Bao,
  Hongjia Zhai, Jialin Liu, Jing Guo, Nan Wang, Siji Pan, Weihong Pan, Weijian
  Xie, Xianbin Liu, Xiaojun Xiang, Xiaoyu Zhang, Xinyu Chen, Yifu Wang, Yipeng
  Chen, Zhenzhou Fan, Zhewen Le, Zhichao Ye, and Ziqiang Zhao.
\newblock {INSPATIO-WORLD}: A real-time {4D} world simulator via spatiotemporal
  autoregressive modeling.
\newblock \emph{arXiv preprint arXiv:2604.07209}, 2026.

\bibitem[Huang et~al.(2025{\natexlab{a}})Huang, Hu, Han, Shi, Tian, He, and
  Jiang]{bib:memory_forcing}
Junchao Huang, Xinting Hu, Boyao Han, Shaoshuai Shi, Zhuotao Tian, Tianyu He,
  and Li~Jiang.
\newblock Memory forcing: Spatio-temporal memory for consistent scene
  generation on {M}inecraft.
\newblock \emph{arXiv preprint arXiv:2510.03198}, 2025{\natexlab{a}}.

\bibitem[Wang et~al.(2026{\natexlab{b}})Wang, Zhang, Cai, Wu, Ackermann, Kuang,
  Zheng, Raji\v{c}, Tang, and Wetzstein]{bib:bullettime}
Yiming Wang, Qihang Zhang, Shengqu Cai, Tong Wu, Jan Ackermann, Zhengfei Kuang,
  Yang Zheng, Frano Raji\v{c}, Siyu Tang, and Gordon Wetzstein.
\newblock {BulletTime}: Decoupled control of time and camera pose for video
  generation.
\newblock In \emph{CVPR}, 2026{\natexlab{b}}.

\bibitem[Huang et~al.(2026{\natexlab{a}})Huang, Jeong, Chen, Gryaditskaya,
  Wang, Lasenby, and Huang]{bib:spacetimepilot}
Zhening Huang, Hyeonho Jeong, Xuelin Chen, Yulia Gryaditskaya, Tuanfeng~Y.
  Wang, Joan Lasenby, and Chun-Hao Huang.
\newblock {SpaceTimePilot}: Generative rendering of dynamic scenes across space
  and time.
\newblock In \emph{CVPR}, 2026{\natexlab{a}}.

\bibitem[Xiao et~al.(2025)Xiao, LAN, Zhou, Ouyang, Yang, Zeng, and
  Pan]{bib:worldmem}
Zeqi Xiao, Yushi LAN, Yifan Zhou, Wenqi Ouyang, Shuai Yang, Yanhong Zeng, and
  Xingang Pan.
\newblock {WorldMem}: Long-term consistent world simulation with memory.
\newblock In \emph{NeurIPS}, 2025.

\bibitem[Li et~al.(2025)Li, Torr, Vedaldi, and Jakab]{bib:vmem}
Runjia Li, Philip Torr, Andrea Vedaldi, and Tomas Jakab.
\newblock {VMem}: Consistent interactive video scene generation with
  surfel-indexed view memory.
\newblock In \emph{ICCV}, 2025.

\bibitem[Wu et~al.(2026)Wu, He, Cheng, Yang, Zhang, Kang, Cai, Wei, Guo, Li,
  and Cheng]{bib:infinite_world}
Ruiqi Wu, Xuanhua He, Meng Cheng, Tianyu Yang, Yong Zhang, Zhuoliang Kang,
  Xunliang Cai, Xiaoming Wei, Chunle Guo, Chongyi Li, and Ming-Ming Cheng.
\newblock {Infinite-World}: Scaling interactive world models to 1000-frame
  horizons via pose-free hierarchical memory.
\newblock In \emph{ICML}, 2026.

\bibitem[Sun et~al.(2026)Sun, Zhang, Wang, Wu, Wang, Wang, Wang, Zhang, Wang,
  and Guo]{bib:hy_worldplay}
Wenqiang Sun, Haiyu Zhang, Haoyuan Wang, Junta Wu, Zehan Wang, Zhenwei Wang,
  Yunhong Wang, Jun Zhang, Tengfei Wang, and Chunchao Guo.
\newblock {WorldPlay}: Towards long-term geometric consistency for real-time
  interactive world modeling.
\newblock In \emph{ICML}, 2026.

\bibitem[{Robbyant Team} et~al.(2026){Robbyant Team}, Gao, Wang, Zeng, Zhu,
  Cheng, Li, Wang, Xu, Ma, Chen, Liu, Cheng, Yao, Zhu, Meng, Zheng, Bai, Chen,
  Shen, Yu, Zhu, Shen, and Ouyang]{bib:lingbot_world}
{Robbyant Team}, Zelin Gao, Qiuyu Wang, Yanhong Zeng, Jiapeng Zhu, Ka~Leong
  Cheng, Yixuan Li, Hanlin Wang, Yinghao Xu, Shuailei Ma, Yihang Chen, Jie Liu,
  Yansong Cheng, Yao Yao, Jiayi Zhu, Yihao Meng, Kecheng Zheng, Qingyan Bai,
  Jingye Chen, Zehong Shen, Yue Yu, Xing Zhu, Yujun Shen, and Hao Ouyang.
\newblock Advancing open-source world models.
\newblock \emph{arXiv preprint arXiv:2601.20540}, 2026.

\bibitem[Huang et~al.(2026{\natexlab{b}})Huang, Ye, Hu, He, Zhang, Shi, Bian,
  and Jiang]{bib:live}
Junchao Huang, Ziyang Ye, Xinting Hu, Tianyu He, Guiyu Zhang, Shaoshuai Shi,
  Jiang Bian, and Li~Jiang.
\newblock {LIVE}: Long-horizon interactive video world modeling.
\newblock In \emph{ICML}, 2026{\natexlab{b}}.

\bibitem[Mao et~al.(2026)Mao, Li, Li, Xu, Ying, and Zhang]{bib:yume15}
Xiaofeng Mao, Zhen Li, Chuanhao Li, Xiaojie Xu, Kaining Ying, and Kaipeng
  Zhang.
\newblock Yume1.5: A text-controlled interactive world generation model.
\newblock In \emph{CVPR}, 2026.

\bibitem[Wei et~al.(2026)Wei, Guo, Li, Xiang, Wei, Zhu, Wang, Wang, Wan, Hou,
  and Fan]{bib:gim}
Zhengxuan Wei, Xu~Guo, Xinghui Li, Xunzhi Xiang, Min Wei, Yiran Zhu, Qiulin
  Wang, Xintao Wang, Pengfei Wan, Xiangwang Hou, and Qi~Fan.
\newblock Geometry-aware implicit memory for video world models.
\newblock \emph{arXiv preprint arXiv:2606.02436}, 2026.

\bibitem[Yu et~al.(2026)Yu, Qian, Li, Wang, Yin, P, Anthony, Ye, Li, Wan, and
  Garg]{bib:mosaicmem}
Wei Yu, Runjia Qian, Yumeng Li, Liquan Wang, Songheng Yin, Sri
  Siddarth~Chakaravarthy P, Dennis Anthony, Yang Ye, Yidi Li, Weiwei Wan, and
  Animesh Garg.
\newblock {MosaicMem}: Hybrid spatial memory for controllable video world
  models.
\newblock \emph{arXiv preprint arXiv:2603.17117}, 2026.

\bibitem[Chen et~al.(2026)Chen, Liang, Zhou, Ding, Liu, Wan, and
  Bai]{bib:hydra}
Kaijin Chen, Dingkang Liang, Xin Zhou, Yikang Ding, Xiaoqiang Liu, Pengfei Wan,
  and Xiang Bai.
\newblock Out of sight but not out of mind: Hybrid memory for dynamic video
  world models.
\newblock \emph{arXiv preprint arXiv:2603.25716}, 2026.

\bibitem[Chen et~al.(2025{\natexlab{b}})Chen, Zhou, Zhao, Wang, Zhang, Huang,
  Sun, Wen, and LI]{bib:flexworld}
Luxi Chen, Zihan Zhou, Min Zhao, Yikai Wang, Ge~Zhang, Wenhao Huang, Hao Sun,
  Ji-Rong Wen, and Chongxuan LI.
\newblock {FlexWorld}: Progressively expanding {3D} scenes for flexible-view
  exploration.
\newblock In \emph{NeurIPS}, 2025{\natexlab{b}}.

\bibitem[Zhang et~al.(2026{\natexlab{a}})Zhang, Le~Moing, Koppula, Rocco,
  Momeni, Xie, Sun, Sukthankar, Barral, Hadsell, Ghahramani, Zisserman, Zhang,
  and Sajjadi]{bib:d4rt}
Chuhan Zhang, Guillaume Le~Moing, Skanda Koppula, Ignacio Rocco, Liliane
  Momeni, Junyu Xie, Shuyang Sun, Rahul Sukthankar, Jo\"elle~K. Barral, Raia
  Hadsell, Zoubin Ghahramani, Andrew Zisserman, Junlin Zhang, and Mehdi S.~M.
  Sajjadi.
\newblock Efficiently reconstructing dynamic scenes one {D4RT} at a time.
\newblock In \emph{CVPR}, 2026{\natexlab{a}}.

\bibitem[Sucar et~al.(2026)Sucar, Insafutdinov, Lai, and Vedaldi]{bib:vdpm}
Edgar Sucar, Eldar Insafutdinov, Zihang Lai, and Andrea Vedaldi.
\newblock {V-DPM}: {4D} video reconstruction with dynamic point maps.
\newblock In \emph{CVPR}, 2026.

\bibitem[Luo et~al.(2026)Luo, Zhou, Lan, Pan, and Loy]{bib:4rc}
Yihang Luo, Shangchen Zhou, Yushi Lan, Xingang Pan, and Chen~Change Loy.
\newblock {4RC}: {4D} reconstruction via conditional querying anytime and
  anywhere.
\newblock In \emph{ICML}, 2026.

\bibitem[Xu et~al.(2025{\natexlab{a}})Xu, Li, Dong, Zhou, Newcombe, and
  Lv]{bib:4dgt}
Zhen Xu, Zhengqin Li, Zhao Dong, Xiaowei Zhou, Richard Newcombe, and Zhaoyang
  Lv.
\newblock {4DGT}: Learning a {4D} {G}aussian transformer using real-world
  monocular videos.
\newblock In \emph{NeurIPS}, 2025{\natexlab{a}}.

\bibitem[Yu et~al.(2025{\natexlab{a}})Yu, Hu, Xing, and
  Shan]{bib:trajectorycrafter}
Mark Yu, Wenbo Hu, Jinbo Xing, and Ying Shan.
\newblock {TrajectoryCrafter}: Redirecting camera trajectory for monocular
  videos via diffusion models.
\newblock In \emph{ICCV}, 2025{\natexlab{a}}.

\bibitem[Lin et~al.(2026)Lin, Liu, Salamanca, Kant, Burgert, Xu, Namekata,
  Zhao, Zhou, Goldblum, Debevec, and Yu]{bib:vista4d}
Kuan~Heng Lin, Zhizheng Liu, Pablo Salamanca, Yash Kant, Ryan Burgert,
  Yuancheng Xu, Koichi Namekata, Yiwei Zhao, Bolei Zhou, Micah Goldblum, Paul
  Debevec, and Ning Yu.
\newblock {Vista4D}: Video reshooting with {4D} point clouds.
\newblock In \emph{CVPR}, 2026.

\bibitem[Aliev et~al.(2020)Aliev, Sevastopolsky, Kolos, Ulyanov, and
  Lempitsky]{bib:npbg}
Kara-Ali Aliev, Artem Sevastopolsky, Maria Kolos, Dmitry Ulyanov, and Victor
  Lempitsky.
\newblock Neural point-based graphics.
\newblock In \emph{ECCV}, 2020.

\bibitem[Wewer et~al.(2024)Wewer, Raj, Ilg, Schiele, and
  Lenssen]{bib:latentsplat}
Christopher Wewer, Kevin Raj, Eddy Ilg, Bernt Schiele, and Jan~Eric Lenssen.
\newblock {latentSplat}: Autoencoding variational {G}aussians for fast
  generalizable {3D} reconstruction.
\newblock In \emph{ECCV}, 2024.

\bibitem[Zhou et~al.(2025)Zhou, Liu, Luo, and
  Huang]{bib:latent_radiance_fields}
Chaoyi Zhou, Xi~Liu, Feng Luo, and Siyu Huang.
\newblock Latent radiance fields with {3D-aware} {2D} representations.
\newblock In \emph{ICLR}, 2025.

\bibitem[Chen et~al.(2024{\natexlab{a}})Chen, Zheng, Xu, Zhuang, Vedaldi, Cham,
  and Cai]{bib:mvsplat360}
Yuedong Chen, Chuanxia Zheng, Haofei Xu, Bohan Zhuang, Andrea Vedaldi, Tat-Jen
  Cham, and Jianfei Cai.
\newblock {MVSplat360}: Feed-forward 360 scene synthesis from sparse views.
\newblock In \emph{NeurIPS}, 2024{\natexlab{a}}.

\bibitem[Hirschorn et~al.(2026)Hirschorn, Sela, Huberman-Spiegelglas, Efrat,
  Alshan, Ideses, Devernay, Zvik, and Fritz]{bib:splatent}
Or~Hirschorn, Omer Sela, Inbar Huberman-Spiegelglas, Netalee Efrat, Eli Alshan,
  Ianir Ideses, Frederic Devernay, Yochai Zvik, and Lior Fritz.
\newblock Splatent: Splatting diffusion latents for novel view synthesis.
\newblock In \emph{CVPR}, 2026.

\bibitem[Lipman et~al.(2023)Lipman, Chen, Ben-Hamu, Nickel, and
  Le]{bib:flow_matching}
Yaron Lipman, Ricky T.~Q. Chen, Heli Ben-Hamu, Maximilian Nickel, and Matt Le.
\newblock Flow matching for generative modeling.
\newblock In \emph{ICLR}, 2023.

\bibitem[{Team Wan} et~al.(2025){Team Wan}, Wang, Ai, Wen, Mao, Xie, Chen, Yu,
  Zhao, Yang, Zeng, Wang, Zhang, Zhou, Wang, Chen, Zhu, Zhao, Yan, Huang, Feng,
  Zhang, Li, Wu, Chu, Feng, Zhang, Sun, Fang, Wang, Gui, Weng, Shen, Lin, Wang,
  Wang, Zhou, Wang, Shen, Yu, Shi, Huang, Xu, Kou, Lv, Li, Liu, Wang, Zhang,
  Huang, Li, Wu, Liu, Pan, Zheng, Hong, Shi, Feng, Jiang, Han, Wu, and
  Liu]{bib:wan}
{Team Wan}, Ang Wang, Baole Ai, Bin Wen, Chaojie Mao, Chen-Wei Xie, Di~Chen,
  Feiwu Yu, Haiming Zhao, Jianxiao Yang, Jianyuan Zeng, Jiayu Wang, Jingfeng
  Zhang, Jingren Zhou, Jinkai Wang, Jixuan Chen, Kai Zhu, Kang Zhao, Keyu Yan,
  Lianghua Huang, Mengyang Feng, Ningyi Zhang, Pandeng Li, Pingyu Wu, Ruihang
  Chu, Ruili Feng, Shiwei Zhang, Siyang Sun, Tao Fang, Tianxing Wang, Tianyi
  Gui, Tingyu Weng, Tong Shen, Wei Lin, Wei Wang, Wei Wang, Wenmeng Zhou, Wente
  Wang, Wenting Shen, Wenyuan Yu, Xianzhong Shi, Xiaoming Huang, Xin Xu, Yan
  Kou, Yangyu Lv, Yifei Li, Yijing Liu, Yiming Wang, Yingya Zhang, Yitong
  Huang, Yong Li, You Wu, Yu~Liu, Yulin Pan, Yun Zheng, Yuntao Hong, Yupeng
  Shi, Yutong Feng, Zeyinzi Jiang, Zhen Han, Zhi-Fan Wu, and Ziyu Liu.
\newblock Wan: Open and advanced large-scale video generative models.
\newblock \emph{arXiv preprint arXiv:2503.20314}, 2025.

\bibitem[Jiang et~al.(2025)Jiang, Han, Mao, Zhang, Pan, and Liu]{bib:vace}
Zeyinzi Jiang, Zhen Han, Chaojie Mao, Jingfeng Zhang, Yulin Pan, and Yu~Liu.
\newblock {VACE}: All-in-one video creation and editing.
\newblock In \emph{ICCV}, 2025.

\bibitem[{Black Forest Labs}(2025)]{bib:flux2}
{Black Forest Labs}.
\newblock {FLUX.2}: Frontier visual intelligence.
\newblock Official model release, 2025.
\newblock URL \url{https://bfl.ai/blog/flux-2}.

\bibitem[Mur-Labadia et~al.(2026)Mur-Labadia, Muckley, Bar, Assran, Sinha,
  Rabbat, LeCun, Ballas, and Bardes]{bib:vjepa21}
Lorenzo Mur-Labadia, Matthew Muckley, Amir Bar, Mido Assran, Koustuv Sinha,
  Mike Rabbat, Yann LeCun, Nicolas Ballas, and Adrien Bardes.
\newblock {V-JEPA} 2.1: Unlocking dense features in video self-supervised
  learning.
\newblock \emph{arXiv preprint arXiv:2603.14482}, 2026.

\bibitem[Wang et~al.(2026{\natexlab{c}})Wang, Yuan, Zheng, Lin, Gao, Chen, Bao,
  Zeng, Zhou, Long, Zhu, Zhang, Cao, and Yao]{bib:spatialvid}
Jiahao Wang, Yufeng Yuan, Rujie Zheng, Youtian Lin, Jian Gao, Lin-Zhuo Chen,
  Yajie Bao, Chang Zeng, Yanxi Zhou, Xiao-Xiao Long, Hao Zhu, Zhaoxiang Zhang,
  Xun Cao, and Yao Yao.
\newblock {SpatialVID}: A large-scale video dataset with spatial annotations.
\newblock In \emph{CVPR}, 2026{\natexlab{c}}.

\bibitem[Zheng et~al.(2025)Zheng, Li, Zhou, and Li]{bib:realcam_vid}
Guangcong Zheng, Teng Li, Xianpan Zhou, and Xi~Li.
\newblock {RealCam-Vid}: High-resolution video dataset with dynamic scenes and
  metric-scale camera movements.
\newblock \emph{arXiv preprint arXiv:2504.08212}, 2025.

\bibitem[Nan et~al.(2025)Nan, Xie, Zhou, Fan, Yang, Chen, Li, Yang, and
  Tai]{bib:openvid}
Kepan Nan, Rui Xie, Penghao Zhou, Tiehan Fan, Zhenheng Yang, Zhijie Chen, Xiang
  Li, Jian Yang, and Ying Tai.
\newblock {OpenVid-1M}: A large-scale high-quality dataset for text-to-video
  generation.
\newblock In \emph{ICLR}, 2025.

\bibitem[{LightX2V Contributors}(2025)]{bib:lightx2v}
{LightX2V Contributors}.
\newblock {LightX2V}: Light video generation inference framework.
\newblock GitHub repository, 2025.
\newblock URL \url{https://github.com/ModelTC/lightx2v}.

\bibitem[Huang et~al.(2024)Huang, He, Yu, Zhang, Si, Jiang, Zhang, Wu, Jin,
  Chanpaisit, Wang, Chen, Wang, Lin, Qiao, and Liu]{bib:vbench}
Ziqi Huang, Yinan He, Jiashuo Yu, Fan Zhang, Chenyang Si, Yuming Jiang, Yuanhan
  Zhang, Tianxing Wu, Qingyang Jin, Nattapol Chanpaisit, Yaohui Wang, Xinyuan
  Chen, Limin Wang, Dahua Lin, Yu~Qiao, and Ziwei Liu.
\newblock {VBench}: Comprehensive benchmark suite for video generative models.
\newblock In \emph{CVPR}, 2024.

\bibitem[Zhou et~al.(2018)Zhou, Tucker, Flynn, Fyffe, and
  Snavely]{bib:realestate10k}
Tinghui Zhou, Richard Tucker, John Flynn, Graham Fyffe, and Noah Snavely.
\newblock Stereo magnification: Learning view synthesis using multiplane
  images.
\newblock \emph{ACM TOG}, 37\penalty0 (4):\penalty0 1--12, 2018.

\bibitem[Duan et~al.(2025)Duan, Yu, Chen, Fei-Fei, and Wu]{bib:worldscore}
Haoyi Duan, Hong-Xing Yu, Sirui Chen, Li~Fei-Fei, and Jiajun Wu.
\newblock {WorldScore}: A unified evaluation benchmark for world generation.
\newblock In \emph{ICCV}, 2025.

\bibitem[Yu et~al.(2025{\natexlab{b}})Yu, Xing, Yuan, Hu, Li, Huang, Gao, Wong,
  Shan, and Tian]{bib:viewcrafter}
Wangbo Yu, Jinbo Xing, Li~Yuan, Wenbo Hu, Xiaoyu Li, Zhipeng Huang, Xiangjun
  Gao, Tien-Tsin Wong, Ying Shan, and Yonghong Tian.
\newblock {ViewCrafter}: Taming video diffusion models for high-fidelity novel
  view synthesis.
\newblock \emph{IEEE TPAMI}, pages 1--18, 2025{\natexlab{b}}.

\bibitem[Huang et~al.(2025{\natexlab{b}})Huang, Zheng, Wang, Liu, Wang, Wu,
  Jiang, Li, Lau, Zuo, and Guo]{bib:voyager}
Tianyu Huang, Wangguandong Zheng, Tengfei Wang, Yuhao Liu, Zhenwei Wang, Junta
  Wu, Jie Jiang, Hui Li, Rynson Lau, Wangmeng Zuo, and Chunchao Guo.
\newblock Voyager: Long-range and world-consistent video diffusion for
  explorable {3D} scene generation.
\newblock \emph{ACM TOG}, 44\penalty0 (6):\penalty0 1--15, 2025{\natexlab{b}}.

\bibitem[Wang et~al.(2004)Wang, Bovik, Sheikh, and Simoncelli]{bib:ssim}
Zhou Wang, Alan~Conrad Bovik, Hamid~Rahim Sheikh, and Eero~P. Simoncelli.
\newblock Image quality assessment: From error visibility to structural
  similarity.
\newblock \emph{IEEE TIP}, 13\penalty0 (4):\penalty0 600--612, 2004.

\bibitem[Zhang et~al.(2018)Zhang, Isola, Efros, Shechtman, and Wang]{bib:lpips}
Richard Zhang, Phillip Isola, Alexei~A. Efros, Eli Shechtman, and Oliver Wang.
\newblock The unreasonable effectiveness of deep features as a perceptual
  metric.
\newblock In \emph{CVPR}, 2018.

\bibitem[Zhang et~al.(2026{\natexlab{b}})Zhang, Zhang, Huang, Tang, Wang, Dai,
  Chen, Li, Li, Chen, Liu, Li, Lyu, and Duan]{bib:mbench}
Shengjun Zhang, Zhang Zhang, Simin Huang, Zhenyu Tang, Hanyang Wang, Chensheng
  Dai, Min Chen, Yifan Li, Yuxin Li, Yingjie Chen, Hao Liu, Chen Li, Jing Lyu,
  and Yueqi Duan.
\newblock {MBench}: A comprehensive benchmark on memory capability for video
  world models.
\newblock \emph{arXiv preprint arXiv:2606.00793}, 2026{\natexlab{b}}.

\bibitem[Moschella et~al.(2023)Moschella, Maiorca, Fumero, Norelli, Locatello,
  and Rodol{\`a}]{bib:latent_analysis}
Luca Moschella, Valentino Maiorca, Marco Fumero, Antonio Norelli, Francesco
  Locatello, and Emanuele Rodol{\`a}.
\newblock Relative representations enable zero-shot latent space communication.
\newblock In \emph{ICLR}, 2023.

\bibitem[Wang et~al.(2025)Wang, Chen, Karaev, Vedaldi, Rupprecht, and
  Novotny]{bib:vggt}
Jianyuan Wang, Minghao Chen, Nikita Karaev, Andrea Vedaldi, Christian
  Rupprecht, and David Novotny.
\newblock {VGGT}: Visual geometry grounded transformer.
\newblock In \emph{CVPR}, 2025.

\bibitem[Oquab et~al.(2024)Oquab, Darcet, Moutakanni, Vo, Szafraniec, Khalidov,
  Fernandez, Haziza, Massa, El-Nouby, Assran, Ballas, Galuba, Howes, Huang, Li,
  Misra, Rabbat, Sharma, Synnaeve, Xu, Jegou, Mairal, Labatut, Joulin, and
  Bojanowski]{bib:dinov2}
Maxime Oquab, Timothée Darcet, Théo Moutakanni, Huy~V. Vo, Marc Szafraniec,
  Vasil Khalidov, Pierre Fernandez, Daniel Haziza, Francisco Massa, Alaaeldin
  El-Nouby, Mido Assran, Nicolas Ballas, Wojciech Galuba, Russell Howes, Po-Yao
  Huang, Shang-Wen Li, Ishan Misra, Michael Rabbat, Vasu Sharma, Gabriel
  Synnaeve, Hu~Xu, Hervé Jegou, Julien Mairal, Patrick Labatut, Armand Joulin,
  and Piotr Bojanowski.
\newblock {DINOv2}: Learning robust visual features without supervision.
\newblock \emph{Trans. Mach. Learn. Res.}, 2024.

\bibitem[He et~al.(2025)He, Xu, Guo, Wetzstein, DAI, Li, and
  Yang]{bib:cameractrl}
Hao He, Yinghao Xu, Yuwei Guo, Gordon Wetzstein, Bo~DAI, Hongsheng Li, and
  Ceyuan Yang.
\newblock {CameraCtrl}: Enabling camera control for video diffusion models.
\newblock In \emph{ICLR}, 2025.

\bibitem[Loshchilov and Hutter(2019)]{bib:adamw}
Ilya Loshchilov and Frank Hutter.
\newblock Decoupled weight decay regularization.
\newblock In \emph{ICLR}, 2019.

\bibitem[Yu et~al.(2024)Yu, Duan, Hur, Sargent, Rubinstein, Freeman, Cole, Sun,
  Snavely, Wu, and Herrmann]{bib:wonderjourney}
Hong-Xing Yu, Haoyi Duan, Junhwa Hur, Kyle Sargent, Michael Rubinstein,
  William~T. Freeman, Forrester Cole, Deqing Sun, Noah Snavely, Jiajun Wu, and
  Charles Herrmann.
\newblock {WonderJourney}: Going from anywhere to everywhere.
\newblock In \emph{CVPR}, 2024.

\bibitem[Engstler et~al.(2025)Engstler, Vedaldi, Laina, and
  Rupprecht]{bib:invisible_stitch}
Paul Engstler, Andrea Vedaldi, Iro Laina, and Christian Rupprecht.
\newblock {Invisible Stitch}: Generating smooth {3D} scenes with depth
  inpainting.
\newblock In \emph{3DV}, 2025.

\bibitem[Yu et~al.(2025{\natexlab{c}})Yu, Duan, Herrmann, Freeman, and
  Wu]{bib:wonderworld}
Hong-Xing Yu, Haoyi Duan, Charles Herrmann, William~T. Freeman, and Jiajun Wu.
\newblock {WonderWorld}: Interactive {3D} scene generation from a single image.
\newblock In \emph{CVPR}, 2025{\natexlab{c}}.

\bibitem[Chen et~al.(2024{\natexlab{b}})Chen, Zhang, Cun, Xia, Wang, Weng, and
  Shan]{bib:videocrafter2}
Haoxin Chen, Yong Zhang, Xiaodong Cun, Menghan Xia, Xintao Wang, Chao Weng, and
  Ying Shan.
\newblock {VideoCrafter2}: Overcoming data limitations for high-quality video
  diffusion models.
\newblock In \emph{CVPR}, 2024{\natexlab{b}}.

\bibitem[Xu et~al.(2025{\natexlab{b}})Xu, Huang, Zou, Chen, Liu, Cheng, Huang,
  and Shi]{bib:easyanimate}
Jiaqi Xu, Kunzhe Huang, Xinyi Zou, Yunkuo Chen, Bo~Liu, Mengli Cheng, Jun
  Huang, and Xing Shi.
\newblock {EasyAnimate}: High-performance video generation framework with
  hybrid windows attention and reward backpropagation.
\newblock In \emph{ACM MM}, 2025{\natexlab{b}}.

\bibitem[Zhou et~al.(2024)Zhou, Wang, Cai, and Yang]{bib:allegro}
Yuan Zhou, Qiuyue Wang, Yuxuan Cai, and Huan Yang.
\newblock {Allegro}: Open the black box of commercial-level video generation
  model.
\newblock \emph{arXiv preprint arXiv:2410.15458}, 2024.

\bibitem[Yang et~al.(2025)Yang, Teng, Zheng, Ding, Huang, Xu, Yang, Hong,
  Zhang, Feng, Yin, Zhang, Wang, Cheng, Xu, Gu, Dong, and Tang]{bib:cogvideox}
Zhuoyi Yang, Jiayan Teng, Wendi Zheng, Ming Ding, Shiyu Huang, Jiazheng Xu,
  Yuanming Yang, Wenyi Hong, Xiaohan Zhang, Guanyu Feng, Da~Yin, Yuxuan Zhang,
  Weihan Wang, Yean Cheng, Bin Xu, Xiaotao Gu, Yuxiao Dong, and Jie Tang.
\newblock {CogVideoX}: Text-to-video diffusion models with an expert
  transformer.
\newblock In \emph{ICLR}, 2025.

\bibitem[Fan et~al.(2025)Fan, Si, Song, Yang, He, Zhuo, Huang, Dong, He, Pan,
  Wang, Jiang, Wang, Gao, Chen, Li, Lin, Qiao, and Liu]{bib:vchitect2}
Weichen Fan, Chenyang Si, Junhao Song, Zhenyu Yang, Yinan He, Long Zhuo, Ziqi
  Huang, Ziyue Dong, Jingwen He, Dongwei Pan, Yi~Wang, Yuming Jiang, Yaohui
  Wang, Peng Gao, Xinyuan Chen, Hengjie Li, Dahua Lin, Yu~Qiao, and Ziwei Liu.
\newblock {Vchitect-2.0}: Parallel transformer for scaling up video diffusion
  models.
\newblock \emph{arXiv preprint arXiv:2501.08453}, 2025.

\bibitem[HaCohen et~al.(2024)HaCohen, Chiprut, Brazowski, Shalem, Moshe,
  Richardson, Levin, Shiran, Zabari, Gordon, Panet, Weissbuch, Kulikov,
  Bitterman, Melumian, and Bibi]{bib:ltx_video}
Yoav HaCohen, Nisan Chiprut, Benny Brazowski, Daniel Shalem, Dudu Moshe, Eitan
  Richardson, Eran Levin, Guy Shiran, Nir Zabari, Ori Gordon, Poriya Panet,
  Sapir Weissbuch, Victor Kulikov, Yaki Bitterman, Zeev Melumian, and Ofir
  Bibi.
\newblock {LTX-Video}: Realtime video latent diffusion.
\newblock \emph{arXiv preprint arXiv:2501.00103}, 2024.

\end{thebibliography}

\clearpage
\appendix
\begingroup
\makeatletter
\renewcommand{\subsection}{\@startsection{subsection}{2}{\z@}
  {8pt plus 2pt minus 1pt}{6pt plus 1pt minus 1pt}{\elvbf}}
\renewcommand{\paragraph}{\@startsection{paragraph}{4}{\z@}
  {6pt plus 2pt minus 1pt}{-1em}{\normalfont\normalsize\bfseries}}
\makeatother

\section{Appendix}
\label{app:appendix}
\suppressfloats[t]

\subsection{4LC Details}
\label{app:implementation-details}

\paragraph{Visual encoders.}\label{app:representation-adaptation}
Source observations are resized and center-cropped to $560\times336$ in the coordinates of the geometry, and each encoder follows the settings in \autoref{tab:representation-settings}.
FLUX.2 VAE~\citep{bib:flux2}, DINOv2~\citep{bib:dinov2}, and VGGT~\citep{bib:vggt} produce one feature map per keyframe, so $\alpha_r(k)=k$ in \autoref{eq:attribute-binding}.
V-JEPA 2.1~\citep{bib:vjepa21} produces one feature step per two frames, so we pair each keyframe with its predecessor, repeating the first keyframe, to obtain one step per keyframe.
Wan2.1 VAE~\citep{bib:wan} encodes the first frame separately and compresses later frames by four; we fill each frame with the most recent keyframe and associate keyframe $k$ at clip index $n_k$ with step $\lceil n_k/4\rceil$.

\begin{table}[tbp]
    \centering
    \caption{\textbf{Visual encoder settings.} $d_r$ is the spatial stride in pixels, $D_r$ the feature dimension, and the target interval the number of frames between readouts.}
    \label{tab:representation-settings}

    \small
    \begin{tabular}{@{}llccc@{}}
        \toprule
        Encoder & Features & $d_r$ & $D_r$ & Target interval \\
        \midrule
        FLUX.2 VAE & Posterior mean of RGB and depth & 8 & 32 & 1 \\
        DINOv2-L & Patch tokens & 14 & 1024 & 1 \\
        VGGT-1B & Aggregator layer-11 tokens & 14 & 2048 & 1 \\
        V-JEPA 2.1 ViT-L & Patch tokens & 16 & 1024 & 2 \\
        Wan2.1 VAE & Latents of RGB and depth & 8 & 16 & 4 \\
        \bottomrule
    \end{tabular}
\end{table}

\paragraph{Motion filtering.}\label{app:motion-filtering}
The motion mask $b_i$ in \autoref{sec:4lc-transformation} marks a unit as dynamic when its estimated motion displaces its projection in the source camera $\Gamma_{k_i}$ by more than $\epsilon$ pixels per frame toward either adjacent observation:
\begin{equation}
    b_i=\mathbf 1\!\left[
    \max_{\pm}
    \frac{\left\|\pi_{\Gamma_{k_i}}\!\left(\mathbf x_i+
    \Delta t_i^\pm\mathbf v_i^\pm\right)-\mathbf p_i\right\|_2}
    {\Delta n_i^\pm}>\epsilon
    \right],
    \label{eq:motion-classification}
\end{equation}
where $\pi_\Gamma$ denotes pixel projection through camera $\Gamma$, and $\Delta t_i^\pm$ and $\Delta n_i^\pm$ are the time and frame gaps to the adjacent observations.
When motion is estimated from multiple observations, units with motion toward neither neighbor are excluded from readout.

\paragraph{Camera projection.}\label{app:camera-readout}
Each unit is assigned to the feature cell nearest to its projection in the target camera.
At each cell, dynamic units from the source time closest to the requested time take priority, and static units fill the remaining cells (\autoref{sec:4lc-transformation}).
Among the remaining candidates, we prefer the source camera closest to the target camera, then the frontmost surface, and finally the unit whose projections lie closest to the source and target cell centers.

\subsection{4LC-Video Details}
\label{app:4lc-video-details}

\paragraph{Condition adapter.}\label{app:generator-alignment}
Wan2.1 latents encode the first frame alone and each later group of four frames, so readouts with target intervals of one, two, and four frames contribute four, two, and one readouts to each later latent step; for 81 frames, they give 81, 41, and 21 readouts over 21 latent steps.
Target-camera rays use Pl\"ucker coordinates~\citep{bib:cameractrl}.

\paragraph{Training.}\label{app:implementation}
The 200K training clips comprise 100K from SpatialVID~\citep{bib:spatialvid}, 40K from RealCam-Vid~\citep{bib:realcam_vid}, and 60K from OpenVid~\citep{bib:openvid}.
We train the condition adapter and control branch for 20K steps on 16 NVIDIA A100 GPUs with a global batch size of 16, using AdamW~\citep{bib:adamw} at a learning rate of $10^{-5}$.

\paragraph{Data augmentation.}\label{app:training-augmentation}
Inspired by NeoVerse~\citep{bib:neoverse}, which simulates degraded conditions from monocular videos, training augmentations vary the available conditions while preserving the target video, times, and cameras.

Each training clip provides $K$ uniformly spaced keyframes, with $K$ drawn from 11 to 21.
To expose the generator to incomplete coverage, we use two geometric augmentations.
Source-frame dropout excludes selected frames from the readout candidates.
Visibility culling retains source units visible from a virtual camera near each source camera, which translates laterally while looking at the scene, yaws, pitches, zooms in, or zooms out.

To account for uncertainty in motion estimation, we perturb the motion threshold $\epsilon$ and construct training conditions with different motion masks.
First-frame sampling restricts the observations to the initial image and constructs geometric conditions from its static state.

Representation and text dropout vary the visual and textual information available to the conditioning path.
Representation dropout removes selected representations and recomputes coverage from the retained ones, and text dropout replaces the text condition with the empty-prompt embedding.
\autoref{tab:training-augmentation} lists the probabilities and ranges.

\begin{table}[tbp]
    \centering
    \caption{\textbf{Training augmentation parameters.} Geometric augmentation applies outside first-frame sampling. Lateral translation is in world units; focal-length factors are relative to the source camera.}
    \label{tab:training-augmentation}

    \small
    \begin{tabular}{@{}llc@{}}
        \toprule
        Type & Parameter & Value \\
        \midrule
        \multirow{6}{*}{Geometric} & Number of keyframes $K$ & 11--21 \\
         & Augmentation probability & 0.6 \\
         & Maximum source-frame drop fraction & 0.3 \\
         & Lateral translation & 0.1--0.3 \\
         & Yaw and pitch & $5^\circ$--$25^\circ$ \\
         & Zoom-in / zoom-out focal-length factor & 1.1--1.3 / 0.7--0.9 \\
        \midrule
        \multirow{2}{*}{Motion} & First-frame sampling probability & 0.1 \\
         & Motion threshold $\epsilon$ (pixels/frame) & 0.05--0.2 \\
        \midrule
        \multirow{3}{*}{\begin{tabular}[c]{@{}l@{}}Representation\\and text\end{tabular}} & Dropout probability, all representations & 0.1 \\
         & Dropout probability, per representation & 0.1 \\
         & Text-dropout probability & 0.1 \\
        \bottomrule
    \end{tabular}
\end{table}

\subsection{Experimental Details}
\label{app:experimental-details}

\paragraph{Spatiotemporal control.}
\label{app:control-evaluation}
As in NeoVerse~\citep{bib:neoverse}, test videos use $K=21$ uniformly spaced keyframes, and its reconstructor estimates camera poses of generated videos.
Each generated and target trajectory is expressed relative to its first frame.
Camera trajectories are pan left, reversed, static, and dolly out at the original scene times; temporal trajectories are reversal, freezing, and pause-and-resume under the source cameras.
Translation error excludes static-camera cases.

\paragraph{Long-term generation.}
\label{app:long-term-protocol}
Each MBench-A~\citep{bib:mbench} case generates 401 frames.
Historical conditions use up to eight uniformly sampled cache entries and their static units ($b_i=0$).
Composition visits them from earliest to latest for Environment cases and from latest to earliest for Causal cases, with a coverage-ratio threshold of 0.95.

\paragraph{Conditioning efficiency.}\label{app:efficiency-protocol}
Acquisition latency and peak GPU memory are measured on an NVIDIA A100, averaging 50 runs after three warmup runs.
Scene construction differs widely across frameworks and cannot be fully aligned: the NeoVerse reconstructor used by NeoVerse and 4LC takes 1.35~s, whereas LiveWorld takes 20.43~s.
For a fair comparison, acquisition and end-to-end timing exclude scene construction, including source reconstruction and encoding, for all frameworks; memory includes weights used for condition acquisition.

\paragraph{Feature readout analysis.}
\label{app:representation-analysis}
For each video, request, and encoder, we uniformly sample 2,048 tokens from the region covered by both acquisition paths.
Relative representations~\citep{bib:latent_analysis} use 128 corresponding anchor features with ten anchor seeds, keeping the statistical tokens fixed across seeds.
We average relative cosine similarities over seeds within each case and give all cases equal weight in \autoref{fig:readouts-cost} (right).
The five requests preserve the source time and camera sequence, reverse scene time, freeze scene time, hold the camera fixed, or reverse the camera trajectory.

\paragraph{Ablations.}
\label{app:ablation-settings}
Each ablation configuration, including the RGB variants, trains its condition adapter and control branch independently on 10K clips for 4K steps on four GPUs with one sample per GPU, and samples with 20 steps.

\subsection{Additional Results}
\label{app:additional-results}

\paragraph{Long-term generation.}\label{app:long-term-qualitative}
\autoref{fig:app-mbench} complements the MBench-A results in \autoref{sec:multi-task-generation} with two cases whose camera trajectory reverses midway and returns to the initial viewpoint.
HY-WorldPlay~\citep{bib:hy_worldplay} regenerates returned regions with shifted layouts and missing structures, and its final frame departs from the first.
4LC-Video adds each generated segment to the cache, and Composition reads out the static units of these entries for the return requests, so return views reproduce the outbound content.
This consistency agrees with its highest Reprojection and \mbox{Environment scores in \autoref{tab:long-horizon-memory}}.

\begin{figure}[tbp]
    \centering
    \includegraphics[width=\linewidth]{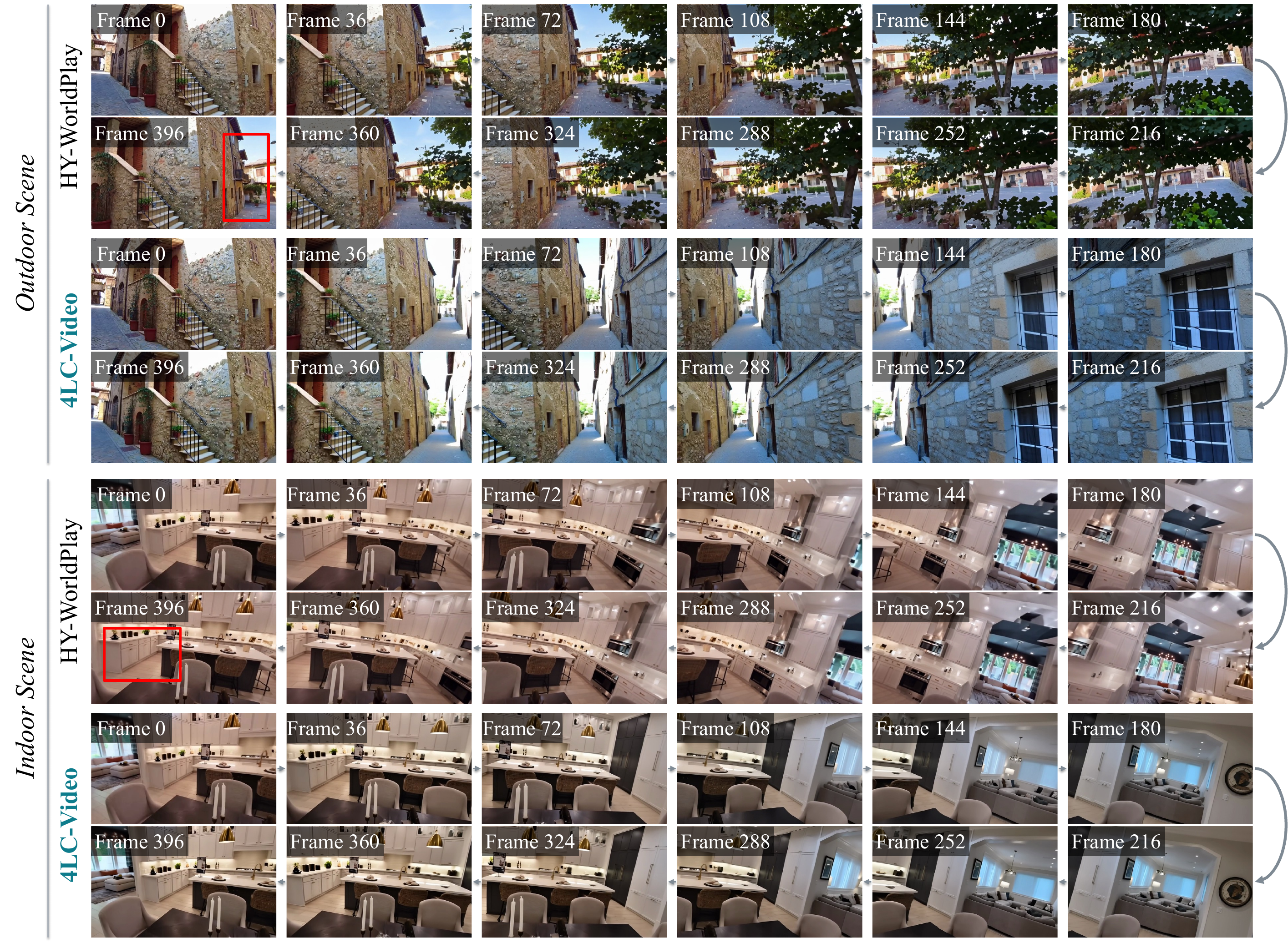}
    \par
    \caption{\textbf{Qualitative comparison for long-term generation on MBench-A.} Each method occupies two rows: the top row follows the outbound trajectory from left to right, and the bottom row follows the return trajectory from right to left, so vertically aligned frames share a camera pose. Red boxes mark structures that change between the first and final frames.}
    \label{fig:app-mbench}
\end{figure}

\paragraph{Generalization to WorldScore.}\label{app:worldscore-generalization}
We evaluate the main model on WorldScore without fine-tuning, using the official pipeline, and compare it with published results~\citep{bib:worldscore} of static scene generators~\citep{bib:wonderjourney,bib:invisible_stitch,bib:wonderworld,bib:voyager} and foundation video models~\citep{bib:videocrafter2,bib:easyanimate,bib:allegro,bib:cogvideox,bib:vchitect2,bib:ltx_video,bib:wan} (\autoref{tab:worldscore-generalization}).
4LC-Video obtains the highest Dynamic score and the best content alignment and style consistency among these methods.
Compared with Wan2.1, it raises camera controllability from 23.53 to 86.87 and 3D consistency from 78.74 to 87.97, and its Static score approaches those of static scene generators.

\begin{table}[tbp]
    \centering
    \caption{\textbf{Quantitative comparison on WorldScore~\citep{bib:worldscore}.} We compare against its published leaderboard results and, following Spatia~\citep{bib:spatia}, report Avg. as the mean of the Static and Dynamic scores; missing motion metrics of static scene generators count as zero.}
    \label{tab:worldscore-generalization}

    \footnotesize
    \begin{tabular}{@{}l*{13}{c}@{}}
        \toprule
        \multirow{2}{*}{Method} & \multicolumn{3}{c}{WorldScore} & \multicolumn{3}{c}{Controllability} & \multicolumn{4}{c}{Quality} & \multicolumn{3}{c}{Dynamics} \\
        \cmidrule(lr){2-4}\cmidrule(lr){5-7}\cmidrule(lr){8-11}\cmidrule(l){12-14}
         & Avg. & Static & Dynamic & Camera & Object & Content & 3D & Photo. & Style & Subj. & Acc. & Mag. & Smooth. \\
        \midrule
        \multicolumn{14}{l}{\textit{Static scene generation}} \\
        WonderJourney & 54.19 & 63.75 & 44.63 & 84.60 & 37.10 & 35.54 & 80.60 & 79.03 & 62.82 & \underline{66.56} & -- & -- & -- \\
        InvisibleStitch & 51.95 & 61.12 & 42.78 & \textbf{93.20} & 36.51 & 29.53 & \textbf{88.51} & \textbf{89.19} & 32.37 & 58.50 & -- & -- & -- \\
        WonderWorld & 61.79 & \underline{72.69} & 50.88 & \underline{92.98} & 51.76 & 71.25 & 86.87 & 85.56 & 70.57 & 49.81 & -- & -- & -- \\
        Voyager & \underline{66.08} & \textbf{77.62} & 54.53 & 85.95 & \textbf{66.92} & 68.92 & 81.56 & 85.99 & \underline{84.89} & \textbf{71.09} & -- & -- & -- \\
        \midrule
        \multicolumn{14}{l}{\textit{Foundation video generation}} \\
        VideoCrafter2 & 50.03 & 52.57 & 47.49 & 28.92 & 39.07 & \underline{72.46} & 65.14 & 61.85 & 43.79 & 56.74 & 47.12 & 30.40 & 29.39 \\
        EasyAnimate & 52.25 & 52.85 & 51.65 & 26.72 & 54.50 & 50.76 & 67.29 & 47.35 & 73.05 & 50.31 & \underline{75.00} & 31.16 & 40.32 \\
        Allegro & 53.64 & 55.31 & 51.97 & 24.84 & \underline{57.47} & 51.48 & 70.50 & 69.89 & 65.60 & 47.41 & 54.39 & \textbf{40.28} & 37.81 \\
        CogVideoX-I2V & 60.64 & 62.15 & \underline{59.12} & 38.27 & 40.07 & 36.73 & 86.21 & 88.12 & 83.22 & 62.44 & 69.56 & 26.42 & \underline{60.15} \\
        Vchitect-2.0 & 40.38 & 42.28 & 38.47 & 26.55 & 49.54 & 65.75 & 41.53 & 42.30 & 25.69 & 44.58 & 33.59 & 33.81 & 21.31 \\
        LTX-Video & 55.99 & 55.44 & 56.54 & 25.06 & 53.41 & 39.73 & 78.41 & \underline{88.92} & 53.50 & 49.08 & \textbf{76.22} & 29.95 & \textbf{71.09} \\
        Wan2.1 & 55.21 & 57.56 & 52.85 & 23.53 & 40.32 & 45.44 & 78.74 & 78.36 & 77.18 & 59.38 & 54.27 & 33.26 & 38.05 \\
        \midrule
        \rowcolor[HTML]{D9E1F2} 4LC-Video (ours) & \textbf{68.46} & 72.18 & \textbf{64.73} & 86.87 & 41.04 & \textbf{76.84} & \underline{87.97} & 68.02 & \textbf{86.41} & 58.10 & 55.66 & \underline{35.78} & 50.65 \\
        \bottomrule
    \end{tabular}
\end{table}

\paragraph{Motion threshold.}\label{app:motion-threshold}
Using the main model on camera trajectories, we compare disabled motion filtering (\autoref{eq:motion-classification}) with filtering at different thresholds (\autoref{tab:motion-threshold}).
Without filtering, estimated motion on static structure enters the conditions as spurious displacement, yielding the largest camera errors and the lowest subject consistency, background consistency, temporal flickering, and motion smoothness.
Training samples the threshold from 0.05 to 0.2 to account for motion-estimation uncertainty (\autoref{tab:training-augmentation}), and performance remains stable from 0.05 to 0.4, including values beyond this range, indicating robustness to the condition changes induced by threshold selection.

\begin{table}[tbp]
    \centering
    \caption{\textbf{Motion filtering and threshold robustness.} The main model is evaluated on the same 100 unseen videos with camera trajectories. $\epsilon$ is in pixels/frame; $\epsilon=0$ disables filtering, and $\epsilon=0.1$ is the default. Bold and red values mark the best and worst results, respectively.}
    \label{tab:motion-threshold}

    \small
    \begin{tabular}{@{}ccccccccc@{}}
        \toprule
        \begin{tabular}[c]{@{}c@{}}Threshold\\$\epsilon$\end{tabular} & $r_{\mathrm{dist}}{\downarrow}$ & $t_{\mathrm{dist}}{\downarrow}$ & \begin{tabular}[c]{@{}c@{}}Subj.\\Consist.\end{tabular} & \begin{tabular}[c]{@{}c@{}}Back.\\Consist.\end{tabular} & \begin{tabular}[c]{@{}c@{}}Temp.\\Flick.\end{tabular} & \begin{tabular}[c]{@{}c@{}}Motion\\Smooth.\end{tabular} & \begin{tabular}[c]{@{}c@{}}Aesth.\\Quality\end{tabular} & \begin{tabular}[c]{@{}c@{}}Imag.\\Quality\end{tabular} \\
        \midrule
        $0\phantom{.00}$ & \textcolor{red}{0.065} & \textcolor{red}{0.138} & \textcolor{red}{89.91} & \textcolor{red}{93.53} & \textcolor{red}{94.44} & \textcolor{red}{97.88} & \textbf{49.08} & \textbf{67.07} \\
        \midrule
        $0.05$ & \textbf{0.035} & 0.092 & 90.61 & 94.06 & 94.81 & 98.03 & 48.50 & 66.88 \\
        $0.1\phantom{0}$ & 0.037 & 0.089 & 90.71 & 94.14 & \textbf{95.47} & \textbf{98.06} & 48.16 & \textcolor{red}{65.49} \\
        $0.2\phantom{0}$ & \textbf{0.035} & \textbf{0.083} & \textbf{90.76} & \textbf{94.24} & 94.88 & 98.04 & 48.15 & 66.61 \\
        $0.4\phantom{0}$ & 0.036 & \textbf{0.083} & 90.74 & 94.23 & 94.88 & 98.04 & \textcolor{red}{48.13} & 66.61 \\
        \bottomrule
    \end{tabular}
\end{table}

\endgroup

\end{document}